\PassOptionsToPackage{table,xcdraw}{xcolor}
\documentclass[acmsmall]{acmart}

\usepackage{longtable}
\usepackage{graphicx}
\usepackage{float}
\usepackage{multirow}
\usepackage{amsmath,amsfonts}
\usepackage{xcolor}
\usepackage{makecell}
\usepackage{soul}
\usepackage[all]{hypcap}
\usepackage[inline]{enumitem}
\usepackage{calc}
\usepackage{tabu}
\usepackage{arydshln}
\usepackage{colortbl}
\usepackage{cellspace}
\usepackage[most, breakable]{tcolorbox}

\graphicspath{{./figures/}}

\newcommand{\numOfSelectedPapers}{109}
\newcommand{\numOfIncludedPapers}{6709}

\newcommand{\numOfVenues}{75}

\newcommand\Tstrut{\rule{0pt}{2.6ex}}        
\newcommand\Bstrut{\rule[-0.9ex]{0pt}{0pt}} 
\tcbset{textmarker/.style={
		enhanced,
		parbox=false,boxrule=0mm,boxsep=0mm,arc=0mm,
		outer arc=0mm,left=6mm,right=3mm,top=7pt,bottom=7pt,
		toptitle=1mm,bottomtitle=1mm,oversize}}
\newtcolorbox{noteBox}{textmarker, breakable,
	borderline west={6pt}{0pt}{green},
	colback=green!10!white}

\setcopyright{none}
\title[Applying Machine Learning in Self-Adaptive Systems]{Applying Machine Learning in Self-Adaptive Systems: A Systematic Literature Review}

\author{Omid Gheibi}
\email{omid.gheibi@kuleuven.be}
\affiliation{
	\institution{Katholieke Universiteit Leuven}
	\streetaddress{Celestijnenlaan 200A}
	\city{Leuven}
	\state{Belgium}
	\postcode{3000}
}
\author{Danny Weyns}
\email{danny.weyns@kuleuven.be}
\affiliation{
	\institution{Katholieke Universiteit Leuven, Linnaeus University}
	\streetaddress{Celestijnenlaan 200A}
	\city{Leuven}
	\state{Belgium}
	\postcode{3000}
}
\author{Federico Quin}
\email{federico.quin@kuleuven.be}
\affiliation{
	\institution{Katholieke Universiteit Leuven}
	\streetaddress{Celestijnenlaan 200A}
	\city{Leuven}
	\state{Belgium}
	\postcode{3000}
}

\begin{document}
	
	\begin{abstract}
		Recently, we witness a rapid increase in the use of machine learning techniques in self-adaptive systems.
		Machine learning has been used for a variety of reasons, ranging from learning a model of the environment of a system during operation to filtering large sets of possible configurations before analysing them. While a body of work on the use of machine learning in self-adaptive systems exists, there is currently no systematic overview of this area. Such overview is important for researchers to understand the state of the art and direct future research efforts.
		This paper reports the results of a systematic literature review that aims at providing such an overview. We focus on self-adaptive systems that are based on a traditional MAPE-based feedback loop (Monitor-Analyze-Plan-Execute). The research questions are centered on the problems that motivate the use of machine learning in self-adaptive systems, the key engineering aspects of learning in self-adaptation, and open challenges in this area. The search resulted in \numOfIncludedPapers{} papers, of which \numOfSelectedPapers{} were retained for data collection. Analysis of the collected data shows that machine learning is mostly used for updating adaptation rules and policies to improve system qualities, and managing resources to better balance qualities and resources. These problems are primarily solved using supervised and interactive learning with classification, regression and reinforcement learning as the dominant methods. 
		Surprisingly, unsupervised learning that naturally fits automation is only applied in a small number of studies. Key open challenges in this area include the performance of learning, managing the effects of learning, and dealing with more complex types of goals. From the insights derived from this systematic literature review we outline an initial design process for applying machine learning in self-adaptive systems that are based on MAPE feedback loops. 
	\end{abstract}


	
	\maketitle

\section{Introduction}
The ever growing complexity of software systems that need to maintain their goals 24/7 while operating under uncertainty motivates the need to equip systems with mechanisms to handle change during operation.  An example is a cloud service that needs to satisfy user performance and minimize operational costs, while operating under dynamically changing workloads. This service may be enhanced with an elasticity module that adjusts resources with changing workload.  

A common approach to handle change is the use of ``internal mechanisms'', such as exceptions (as a feature of a programming language) and fault-tolerant protocols. The application of such mechanisms is often domain-specific and tightly bound to the code. This makes it costly to build, modify, and reuse solutions~\cite{garlan2004rainbow}. 
In contrast, change can be handled using ``external mechanisms'' that are based on the concept of a feedback loop. An important paradigm in this context is self-organization, where relatively simple elements apply local rules to adapt their interactions with other elements in response to changing conditions in order to cooperatively realize the system goals~\cite{Heylighen2002,parunak_brueckner_2015}. Another important paradigm is control-based adaptation that relies on the mathematical basis of control theory for designing feedback loop systems and analyzing and guaranteeing key properties~\cite{Hellerstein04,7929422}. 

In this paper, we focus on architecture-based adaptation, which is an extensively studied and applied approach to handle change~\cite{kephart2003vision,garlan2004rainbow,weyns2012forms,ModelsAtRuntime,Roadmap2009,Weyns19}. Architecture-based adaptation 
relies on a feedback loop that monitors the system and its context and adapts the system to ensure its goals, or degrade gracefully if necessary. 
Pivotal in tackling this task is the use of 
runtime models~\cite{garlan2004rainbow,ModelsAtRuntime} that enable the system to reason about (system-wide) change and make adaptation decisions. The feedback loop localizes the adaptation concerns in separable system elements that can be analyzed, modified, and reused across systems. 
Example practical applications of architecture-based adaptation are management of renewable energy production plants~\cite{CAMARA2016507}, information systems for public administration~\cite{SilvaSPC17}, 
and automation of the management of Internet of Things applications~\cite{978-3-030-00761}.

Realizing feedback loops for architecture-based adaptation is in general not a trivial task. 
Over the past years, several techniques have been investigated to support the design and operation of self-adaptation. One of these techniques is search-based software engineering. For instance,~\cite{Cheng2013} argues for the use of evolutionary computation to generate and analyze models of dynamically adaptive systems in order to deal with uncertainties both at development time and runtime. Our focus in this paper is on another prominent line of research that applies  machine learning techniques in the design and operation of self-adaptation.  
We highlight a few of the incentives to apply machine learning techniques to architecture-based self-adaptive systems.
Online verification of rigorously specified runtime models enables providing guarantees about the adaptation decisions. However, formal verification of runtime models for all possible adaptation options during operation can be time consuming. We may then use a learning mechanism to filter the configurations before starting analysis. Another challenging aspect is the design of runtime models of complex software systems. These models may become particularly complicated up to the level that it may be infeasible to design the models if the structure of the system or its context is not known beforehand. Hence, the models need to be derived during operation, for which we may use learning techniques.

At the current point in time we have a sizeable body of work in this area. Yet, even though the amount of research that has been conducted on this topic is quite substantial, there is no clear view on the state of the art. Such an overview is important for researchers in this area as it will document the current body of knowledge and clarify open challenges. 
To tackle this problem, we performed a systematic literature review~\cite{keele2007guidelines}. The goal of this study is to provide a systematic overview of the state of the art on the application of machine learning methods in self-adaptation. 
The survey is centered on (i) the problems that motivate the use of machine learning in self-adaptive systems, (ii) key engineering aspects of learning applied in self-adaptation, and (iii) open challenges. 

The remainder of this paper is structured as follows.  Section~\ref{sec:focus} starts with explaining background and outlining the focus of the literature review.  
In Section~\ref{sec:relatedWork}, we position this review in the landscape of related review work. 
Section~\ref{sec:protocolDefinition} then summarizes the protocol of the study that includes the research questions, the search string, inclusion and exclusion criteria, the data items that we extracted from the papers, and the methods we used for the analysis.  Section~\ref{sec:results} reports the results from the analysis of the collected data, providing answers to the research questions. 
In Section~\ref{sec:discussion}, from the main findings of the literature review, we present an initial design process for applying machine learning in self-adaptive systems, we present opportunities for further research, and we discuss threats to validity. Finally, we wrap up and conclude the paper in Section~\ref{sec:conclusion}. 

\section{Background and Review Focus}\label{sec:focus}
In this section, we briefly introduce self-adaptive software with a MAPE-based feedback loop, and we summarize the essential dimensions of machine learning methods. Then we clarify the distinction between an adaptation problem and a learning problem in the context of this study, and we highlight the importance of the use of machine learning in self-adaptive systems.

\subsection{MAPE-based Self-adaptation}\label{subsec:MAPE}

This study focuses on self-adaptive systems based on architecture-based adaptation~\cite{garlan2004rainbow,Kramer2007SMS, weyns2012forms}. 
Such a self-adaptive system comprises a \textit{managed system} that is controllable and subject to adaptation, and the \textit{managing system} that performs the adaptations of the managed system. The managed system operates in an \textit{environment} that is non-controllable. The managing system realizes a feedback loop that comprises four essential functions: Monitor-Analyze-Plan-Execute that share Knowledge; often referred as MAPE-K or MAPE in short~\cite{kephart2003vision}. The monitor tracks the managed system and the environment in which the system operates and updates the knowledge. The analyzer uses the up-to-date knowledge to evaluate the need for adaptation, possibly exploiting rigorous analysis techniques~\cite{Calinescu11,ActivFORMS,Camara2016SMGs} or simulations of runtime models~\cite{7573167}. Such analysis may apply rigorous methods to provide guarantees for the If adaptation is required, it analyzes alternative configurations of the managed system.  We refer to these alternative configurations as the adaptation options. The planner then selects the best option based on the adaptation goals and generates a plan to adapt the system from its current configuration to the new configuration. Finally, the executor executes the adaptation actions of the plan. It is important to highlight that MAPE provides a reference model that describes a managing system's essential functions and the interactions between them. A concrete architecture maps the functions to corresponding components, which can be a one-to-one mapping or any other mapping, such as a mapping of the analysis and planning functions to one integrated decision-making component. 

In this literature review, we consider studies that are based on the MAPE reference model that maps the MAPE functions (or some of them) to a specific component-based architecture.

\subsection{Machine Learning}\label{subsec:ML}
T.~Mitchell defines machine learning as follows: ``a computer program is said to learn from experience $E$ concerning some class of tasks $T$ and performance measure $P$, if its performance at tasks in $T$, as measured by $P$, improves with experience $E$''~\cite{mitchell1997machine}. 
For example, consider a self-adaptive sensor network that needs to keep packet loss and latency under given thresholds. The training experience ($E$) from which we learn could be the results of the analysis of adaptation options. The task ($T$) could be a classification of the adaptation options in two classes: those that are predicted to comply with the goals (and should be analyzed) and those that are predicted not to comply (and should not be analyzed). The performance measure ($P$) to perform this task could be the comparison of predicted values of packet loss and latency with the threshold values of the respective adaptation goals. In this example, learning (classification) supports the analysis stage of the feedback loop by reducing a large number of adaptation options, aiming to improve the efficiency of analysis.

In the field of machine learning, a distinction can be made in four dimensions:~\cite{shalev2014understanding,bishop2006pattern}:
\begin{itemize}
	\item \textit{Unsupervised vs Supervised vs Interactive}: 
	An unsupervised learner aims at finding previously unknown patterns in data sets without preexisting labels. One of the main methods used in unsupervised learning is cluster analysis. Cluster analysis identifies commonalities in the data and reacts based on the presence or absence of such commonalities in new data. A supervised learner learns a function that maps an input to an output based on example input-output pairs. The function is inferred from labeled data and can then be used to map new data. An interactive learner collects the input-output pairs by interaction with the environment. The learner uses a learning model to make predictions that  are then used to perform actions in the environment. The environment then provides feedback on the actions that the learner incorporates  in its learning model improving the learning process; a classic example is reinforcement learning. We refer to the basic dimension that distinguishes unsupervised, supervised, and interactive learning as \textit{learning type}.  
	
	\item \textit{Active vs Passive}: In active learning a learning algorithm interactively queries some information source in the environment (e.g., a user or teacher) to obtain the desired outputs at new data points. These outputs in turn are used to affect the environment. A passive learner only perceives the information from the environment without affecting it.
	
	\item \textit{Adversarial vs Non Adversarial}: Adversarial learning attempts to fool models through malicious input. This technique can be applied to attack standard learning models. An example is an attack in spam filtering, where spam messages are obfuscated through the manipulation of the text. Non adversarial learning has no concept of malicious input.

	\item \textit{Online vs Batch Protocol}: In online learning data becomes available in a sequential order and is used to update the learning model for future data at each step. Batch learning on the other hand generates the learning model by learning on the entire training data set at once.
	
\end{itemize}

In the literature review, we consider the four dimensions of machine learning methods. 

\subsection{Adaptation Problem versus Learning Problem}\label{subsec:AP-LP}

\begin{figure}[htbp]
	\centering
	\includegraphics[width=0.85\linewidth]{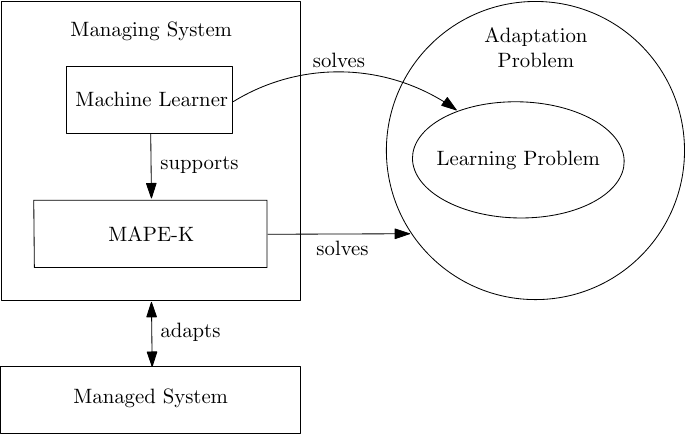}
	\caption{Relation of learning problem and adaptation problem with the components of the managing system.}
	\label{fig:LP-AP}
\end{figure}

As explained above, in this survey we target software systems that comprise of two parts: a managed system that is adapted by a managing system, as illustrated in Figure~\ref{fig:LP-AP}. The managing system is based on a MAPE-K feedback loop that solves an \textit{adaptation problem}. An adaptation problem relates to one or more concerns of a managed system that typically pertain to quality properties. The MAPE-K feedback loop is supported by a \textit{machine learner} that solves a particular \textit{learning problem}. Hence, in this study, we look at learning problems that are part of adaptation problems. 

Consider the example we used in the introduction of a cloud service as a managed system. The adaptation problem is to ensure user performance while minimizing operational cost for the owner under changing workloads. To that end, the cloud service is extended with an elasticity module. This module realizes a feedback loop that dynamically adjusts the resources of the cloud service based on changing workloads. A learning problem in this context may be the prediction of the workload of the cloud service. Such learner would support the monitor and analyzer of the feedback loop of the elasticity module. Hence, the elasticity of the cloud service is realized by the collaboration of the MAPE-K feedback loop and the machine learner that together form the managing system. 

\subsection{Importance of Machine Learning in Self-Adaptive Systems}\label{subsec:importance}

A recent book structures the field of self-adaptation in seven ``waves'' that highlight the important research areas of the past two decades that have contributed to the current body of work~\cite{weyns2020book}. The seventh wave focuses on machine learning techniques as a means to enhance the realization of a self-adaptive software system. The book argues that machine learning can be used to support different activities of the MAPE workflow of self-adaptive systems and highlights three characteristic use cases. 
The first use case enhances the monitor function of a self-adaptive system with a Bayesian estimator that keeps a runtime model up to date. A concrete example is described in~\cite{Epifani:2009}. This simple use case underpins the power of machine learning when dealing with parametric uncertainties represented in runtime models. The second use case enhances the analyzer function with a classifier that enables large sets of adaptation options to be reduced at runtime, improving the efficiency of the analysis phase of self-adaptation. A concrete example is described in~\cite{Quin2019}. This use case shows how learning can help to deal with the complexity that comes with the increasing scale of self-adaptive systems. Finally, the third use case enhances various feedback loop functions with a learning strategy that combines fuzzy control and fuzzy Q-learning to adjust and improve auto-scaling rules of a cloud infrastructure at runtime. An example is described in~\cite{7515437}. This use case shows how machine learning can help to support decision making in self-adaptive systems that are subject to complex types of uncertainties. 

These examples underpin the importance of machine learning techniques in self-adaptive systems. The aim of this paper is to study the use of machine learning in self-adaptation in a systematic way in order to document the current body of knowledge in this area and identify open challenges.

\section{Related Reviews}\label{sec:relatedWork}
A number of reviews related to this study have been published, but they differ in the methodology used, the domain studied, or the types of feedback loops considered. We highlight key aspects of these studies and position our work to these reviews. Table~\ref{tab:related-reviews} summarizes the related work. 

Klaine et al.~\cite{Klaine2017} performed a review on machine learning in self-organizing networks. The authors classified papers based on their learning solution and self-organizing use-case. Moreover, they provided a general guideline to deciding which machine learning algorithm is proper for which use-case in self-organizing networks. That work has three main differences compared to our systematic literature review: 
\begin{enumerate*}
	\item methodology: their work is a basic review without any specific query, data extraction and analysis phase, in contrast to our study, which is a systematic literature review; 
	\item domain: their work focused on a specific domain, i.e., self-organizing networks, in contrast to our work that does not put any constraint on the application domain in self-adaptive systems; 
	\item managing system: their work mainly focused on control-theoretic feedback loops, in contrast to our work that focuses on self-adaptive systems with MAPE-based feedback loops. 
\end{enumerate*}

D'Angelo et al.~\cite{D'Angelo2019} presented a three-dimensional framework to categorize and compare work on learning capabilities in collective self-adaptive systems. This framework approached the applied learning methods from three dimensions:  autonomy, knowledge access, and behavior. Although that work is analogous with our work in terms of using machine learning in self-adaptive systems,  the main differences are: 
\begin{enumerate*}
	\item research methodology: they followed the principles of a literature review but with a more specific focus than our study; 
	\item domain: they focused on decision-making realized by so called collective self-adaptive systems, while our work goes beyond decentralized decision-making and considers all types of MAPE-based self-adaptive systems that are selected via inclusion/exclusion criteria. 
	\item feedback loops: they focused on decentralized self-adaptive systems or multi-agent systems, while our work considers all types of MAPE-based self-adaptive systems. 
\end{enumerate*}

Gambi et al.~\cite{Gambi2013} studied self-adaptive controllers applied in the Cloud. The authors focus on two main dimensions: adaptability (flexibility and scope of the adaptation of the system), and reliability (accuracy of the assurances for reliability). Based on these dimensions, the authors categorized existing control methods in three groups: 
\begin{enumerate*}
	\item controllers that prefer adaptability over reliability;
	\item controllers that prefer reliability over adaptability;
	\item and controllers that balance between the two. 
\end{enumerate*}
The main differences with our work are:
\begin{enumerate*}
	\item research methodology: their work is a basic literature study, while our study follows the guidelines of a systematic literature review; 
	\item domain: their work is focused on the Cloud domain, while we consider all application domains. Moreover, their work focused on specific concerns in Cloud without a particular focus on learning methods. In contrast, our work aims at identifying existing issues in a variety of domains that have been tackled by machine learning. 
	\item managing system: the proposed work has no particular constraints in terms of control approaches used, while our work is scoped on MAPE-based feedback loop systems.
\end{enumerate*}

Lorido-Botran et al.~\cite{lorido2014} classified techniques used for auto-scaling of Cloud environments. The authors identified five groups of techniques: threshold-based rules, reinforcement learning, queuing theory, control theory, and time series analysis. The main differences with our work are:
\begin{enumerate*}
	\item research methodology: the authors do not report any specific research methodology they applied in their work. In contrast we applied a systematic literature review; 
	\item domain: their work focused on a specific concern (auto-scaling) within a particular application domain (Cloud). In contrast, problems and domains are open in our study, and the focus is on the use of learning methods to support self-adaptation;
	\item feedback loops: there is no limitation on controlling approaches in their work, opposite to our criteria to focus on MAPE-based feedback loops.
\end{enumerate*}

Another survey studied the classification of workload prediction methods in cloud computing~\cite{masdari2019}. The motivation for this review is the crucial role of the workload prediction in auto-scaling and resource management of clouds. As a result, the authors identified 11 classes of prediction methods, including regression-based, classifier-based, stochastic-based, and wavelet-based methods.
The main differences with our work are:
\begin{enumerate*}
	\item research methodology: in contrast to our systematic literature review, there is no specific research methodology applied in the presented work;
	\item domain: the presented work focuses on specific methods to solve specific problems in the Cloud, whereas we target a broad range of problems and application domains. 
	\item  feedback loops: they focus on general prediction methods without assuming any particular system structure, while our work is scoped on utilizing machine learning techniques in architecture-based self-adaptive systems.
\end{enumerate*}

Cui et al.~\cite{cui2018} performed a survey on the application of machine learning in the IoT domain, e.g., for security, traffic profiling, and device identification in IoT. Two main classes of application domains that they identified are personal health and industrial applications. However, our work is different from their work in the following terms:
\begin{enumerate*}
	\item research methodology: their work is a basic literature study, while we studied the literature using a systematic method; 
	\item domain: their work focused on the domain of IoT, while we did not put any constraints in our study on application domains;  
	\item feedback loops: their work is not limited to self-adaptive systems, although the work is helpful for self-adaptation. Hence, in their work, the nature of the managing system does not matter, in contrast to our work that targets MAPE-based feedback loop systems. 
\end{enumerate*}

Finally, Saputri and Lee~\cite{9249012} performed a literature review that aimed at helping software engineers in proper selection of machine learning techniques based on adaptation concerns at hand. 
Based on the results, the authors proposed an initial taxonomy for choosing machine learning techniques in self-adaptation. The results of this study are difficult to interpret as the authors mix learning types, learning tasks, and learning methods. Similarly, 
the authors used a concept called ``concern objects for adaptation in self-adaptive systems'' that ``refers to the concerns in the adaptation that are addressed using machine learning approaches''. These ``concerns objects'' are architecture, behavior, framework, model, and verification. It is not clear why these concepts were chosen and neither is it clear how a classification can be done based on these concepts as they have obvious overlaps. Some of the other differences compared to our study are: 
\begin{enumerate*}
	\item methodology: the authors performed a systematic literature review, yet, it is remarkable that the automatic search resulted in only 315 papers (78 selected) compared to over 6700 papers in our study (109 selected); 
	\item domain: the paper states that the focus of their work is self-adaptation, yet, a variety of papers are included on agents, self-organization, and general software systems;
	\item feedback loops: the paper states that it focuses on self-adaptation, yet, the authors report that 55 of the 87 selected papers do not have any concept of MAPE; the term feedback loop is not mentioned in the paper. 
\end{enumerate*}

\small
\begin{table}[h!]
	\caption{Summary of related reviews (SLR refers to Systematic Literature Review).}\label{tab:related-reviews}
	\vspace{-5pt}
	\begin{tabular}{l|l|l|l|}
		\hline
		\multicolumn{1}{|c|}{\textbf{Related review}} & \multicolumn{1}{c|}{\textbf{Method}} & \multicolumn{1}{c|}{\textbf{Domain}} & \multicolumn{1}{c|}{\textbf{Type feedback loop}} \\ \hline
		\multicolumn{1}{|l|}{Klaine et al.~\cite{Klaine2017}} & 
		Literature review                   & 
		\begin{tabular}[c]{@{}l@{}}Self-organizing networks\end{tabular} &  
		Control-theoretic
		\\ \hline
		\multicolumn{1}{|l|}{D'Angelo et al.~\cite{D'Angelo2019}} & SLR & Collective self-adaptive systems                          &                                    
		\begin{tabular}[c]{@{}l@{}}Decentralized\end{tabular}
		\\ \hline
		\multicolumn{1}{|l|}{Gambi et al.~\cite{Gambi2013}} &
		Not specified&
		\begin{tabular}[c]{@{}l@{}}Cloud\end{tabular}&
		Not constrained
		\\ \hline
		\multicolumn{1}{|l|}{Lorido-Botran et al.~\cite{lorido2014}} &
		Not specified&
		\begin{tabular}[c]{@{}l@{}} Cloud\end{tabular}&
		Not constrained
		\\ \hline
		\multicolumn{1}{|l|}{Masdari et al.~\cite{masdari2019}} &
		Not specified&
		\begin{tabular}[c]{@{}l@{}} Cloud\end{tabular}&
		Systems in general 
		\\ \hline
		\multicolumn{1}{|l|}{Cui et al.~\cite{cui2018}} &
		Literature review&
		\begin{tabular}[c]{@{}l@{}} IoT\end{tabular}&
		Systems in general 
		\\ \hline
		\multicolumn{1}{|l|}{Saputri and Lee~\cite{9249012}} &
		SLR&
		\begin{tabular}[c]{@{}l@{}} Self-adaptive systems \end{tabular}&
		Systems in general 
		\\ \hline
	\end{tabular}
	\vspace{-5pt}
\end{table}
\normalsize

\section{Summary Protocol}\label{sec:protocolDefinition}

This study used the methodology of a systematic literature review as described in~\cite{keele2007guidelines}. The methodology defines the way in which a literature review should be performed so that the relevant papers are properly identified, evaluated, and interpreted. A systematic literature review is composed of three stages: planning, execution, and reporting. During the planning stage, a protocol is defined for the study. This protocol includes the research questions of the study, the sources to search for papers, the search string to collect papers, inclusion and exclusion criteria to select relevant papers, and the data items that need to be collected from the selected papers to answer the research questions. In the execution phase the search string is applied, papers are collected, and the data is extracted. In the reporting phase the collected data is analyzed and interpreted, the research questions are answered, useful insights are documented, and potential threats to the validity of the study are discussed. 

We conducted the systematic literature review with three researchers that jointly developed the protocol. 
One researcher performed the automatic search, resulting in 6709 papers.  To avoid bias when selecting papers, i.e., the primary studies, we applied the inclusion and exclusion criteria as follows. For a first batch of 196 randomly identified papers two researchers selected papers. The results were then compared and in case of differences, the two researchers resolved any conflicts. If no consensus could be reached, the third researcher was involved to make a decision in consensus. We repeated this process until the two reviewers selected the same papers. Concretely, we used three more rounds, where the researchers applied the inclusion/exclusion criteria respectively to 111, 66 and 47 randomly selected papers. The number of conflicts decreased to zero in the final round. One researcher then applied the inclusion/exclusion criteria to the remaining papers. We used a similar process for the extraction of the data. We applied two rounds of data extraction, in each round two researchers extracted data of five papers. One researcher then extracted the data of the remaining papers and a sample of five other papers were crosschecked by the two other researchers. Finally, the analysis and reporting was jointly done by the three researchers (we explain the process we used in Section~\ref{subsec:analysis_data}).  

We now briefly explain the main parts of the protocol. The full description of the protocol together with a replication package is available at the study website.\footnote{\url{https://people.cs.kuleuven.be/danny.weyns/material/ML4SAS/SLR/}}

\subsection{Research Questions}
We formulated the goal of the study using the classic Goal-Question-Metric (GQM) approach~\cite{van2002goal}:
\begin{quote}
	\textit{Purpose}: analyze and characterize\\
	\textit{Issue}: the use of machine learning\\
	\textit{Object}: to support self-adaptation based on MAPE-K feedback loops\\
	\textit{Viewpoint}: from a researcher’s viewpoint.
\end{quote}

We translated the overall goal of the review to three concrete research questions:

\begin{description}[leftmargin=!, labelwidth=\widthof{\bfseries RQ4:}]
	\item[RQ1:] What problems have been tackled by machine learning in self-adaptive systems?
	\item[RQ2:] What are the key engineering aspects considered when applying learning in self-adaptation?  
	\item[RQ3:] What are open challenges for using machine learning in self-adaptive systems?
\end{description} 

With RQ1, we wanted to gain insight into the motivations why machine learning has been applied in self-adaptation and in particular for what problems machine learning has been used in self-adaptive systems. 
With RQ2, we wanted to understand key aspects in the realization of self-adaptation. This included the MAPE functions of feedback loops that are supported by learning, the learning methods that have been used, and the representation of learning dimensions of these methods (as explained in the background section). With RQ3, we wanted to get insight into the limitations and the challenges in applying machine learning in self-adaptation that are reported in the studies.

\subsection{Searched Sources and Search Query}

The search strategy combined automatic with manual search. In a first step we performed an automated search on the three main data search engines where research results on self-adaptation are published: IEEE Explore, ACM Digital Library, and Springer Link. 

To identify the search string and ensure that it finds all the relevant papers, we applied pilot searches for two main venues that publish research on self-adaptation, looking at the years 2017 till 2019: TAAS and SEAMS. During these pilots we combined different keywords and compared the results of the query with manually identified papers. This way, we iteratively adjusted the search query such that the automatic search found all the manually identified papers with a minimum number of false positives. This resulted in the following search string that we used to select papers based on title and abstract:

\vspace{5pt}
\begin{center}
	\mbox{\ \ \ \ \ \ \ \ \ \ \ \ }\textit{\small(Title:learn*~AND~(Title:adaptation OR Title:self*))} \textit{OR}\\
	\mbox{\ \ \ \ \ \ }\textit{(Abstract:learn* AND (Abstract:adaptation OR Abstract:self*))}
\end{center}
\vspace{5pt}

\noindent
After finalizing the search string, we launched a full automated search to collect the papers (i.e., all the primary studies).\footnote{ Note that we instantiated the search string according to the interface specific to each of the digital libraries we used. For instance, for the ACM library (https://dl.acm.org/search/advanced), the search query was formulated as ``Title:(learn* AND (self* OR "adaptation")) OR Abstract:(learn* AND (self* OR "adaptation"))'' and the publication date was set ``From Jan. 2003 To May 2020''. For a detailed description of the settings and process we used for each library, we refer to the study website. } We then manually applied the inclusion and exclusion criteria to these papers to select relevant papers.

\subsection{Inclusion and Exclusion Criteria}
We used the following inclusion criteria to select papers:

\begin{itemize}
	\item Papers that were published between January 2003 to May 2020\footnote{We selected 2003 as start date similar to previous reviews, based on the emergence of venues at the time that were dedicated to adaptation such as the International Conference on Autonomic Computing (ICAC).} 
	\item Papers that used machine learning methods in self-adaptive software systems that are based on a MAPE feedback loop. We target adaptation of application software or services that support the application software (in contrast to direct adaptation of hardware). 
	\item Papers that provided a basic level of evaluation of the research, which may be in the form of a simple evaluation of application scenarios, a systematic simulation of a system, rigorous analysis, empirical evaluation, up to a real-world case study.
\end{itemize}

\noindent We used the following exclusion criteria:

\begin{itemize}
	\item Surveys and roadmap papers, as we were only interested in studies that concretely apply machine learning in self-adaptive systems with a minimum level of evaluation.
	\item Tutorials, short papers\footnote{Papers with less than 5 pages are excluded.}, editorials etc. because these papers do not provide sufficient data for the review.
\end{itemize}

A paper was selected if it met all inclusion criteria and did not meet any exclusion criterion.

\subsection{Data Items}
To answer the research questions, we defined a set of data items to be extracted from the papers. Table~\ref{tab:data_items} gives an overview of the data items that we briefly discuss now. 
The concrete options for each data item are further discussed in the next section. For a detailed description of the data items, we refer to the protocol that is available at the website of the systematic literature review.

\begin{table}[htbp]
	\centering
	\caption{Data extraction items}
	\label{tab:data_items}\vspace{-5pt}
	\begin{tabular}{|l|l|l|}
		\hline
		\textbf{Item ID} & \textbf{Item} & \textbf{Use} \\ \hline
		F1      & Authors             & Documentation  \\ \hline
		F2      & Year                & Documentation  \\ \hline
		F3      & Title               & Documentation  \\ \hline
		F4      & Venue               & Documentation  \\ \hline
		F5      & Citation count      & Documentation  \\ \hline
		F6      & Quality score       & RQ1-3          \\ \hline
		F7      & Adaptation problem  & RQ1            \\ \hline
		F8      & Learning problem    & RQ1            \\ \hline 
		F9      & MAPE function(s) supported by learning     & RQ2       \\ \hline
		F10     & Dimensions of learning methods    & RQ2       \\ \hline
		F11      & Learning method(s) used to support  self-adaptation & RQ2            \\ \hline 
		F12     & Application domain  & RQ1            \\ \hline
		F13     & Limitations & RQ3       \\ \hline
		F14     & Challenges & RQ3        \\ \hline

	\end{tabular}
	\vspace{5pt}
\end{table}

\begin{description}[leftmargin=!, labelwidth=\widthof{\bfseries F1-5:}]
	\item[F1-5:] The data items author(s), year, title, venue, citation count are used for documentation purpose.
	
	\item[F6:] Quality score assesses the quality of the reporting of the research in the paper, which is important for data analysis and interpretation of the results. Inline with~\cite{dybaa2008empirical} and~\cite{7929422}, we assessed the following quality items: 
	\begin{enumerate*}[label=(\arabic*)]
		\item problem definition of the study, 
		\item problem context, i.e., the way the study is related to other work, 
		\item research design, i.e., the way the study was organized,
		\item contributions and study results, 
		\item insights derived from the study, 
		\item limitations of the study.
	\end{enumerate*}
	For each item, we consider three quality levels: explicit description (2 points), general description (1 point), and no description (0 points). We calculate a quality assessment score (max 12) as the sum of the scores of all items of a study.
	
	\item[F7:] The adaptation problem that is tackled by the managing system. Here we look at what is the main problem for which self-adaptation is applied. The options are collected during data-gathering. 
	
	\item[F8:] The problem that is tackled by machine learning in the realization of self-adaptation. Here we look at the motivations why machine learning is used to support self-adaptation, i.e., what concrete problem is solved. The options are collected during data-gathering. 
	
	\item[F9:] The MAPE functions supported by machine learning. Options are: monitor, analyzer, planner, executor, and any combination of the four basic functions. 

	\item[F10:] The dimensions of the learning methods applied (as explained in  Section~\ref{subsec:ML}). Options for the dimensions are : unsupervised-interactive-supervised, active-passive, adversarial-non adversarial, and online-batch.
	
	\item[F11:] The concrete machine learning methods used to solve learning problems, e.g., linear regression, support vector machine, reinforcement learning, classical neural network, deep neural network, etc. The concrete options are collected during data-gathering. 
	
	\item[F12:] The application domain in which the machine learning method is applied to support self-adaptation. Initial options are: service-based system, cyber-physical system, internet-of-things, robotics, and cloud. Additional options are collected during data-gathering. 
	
	\item[F13:] The limitations reported in the paper. The options are collected during data-gathering. 
	
	\item[F14:] The challenges for future research on machine learning for self-adaptation reported in the paper. The options are collected during data-gathering. 
	
\end{description}

\subsection{Approach for Analysis}\label{subsec:analysis_data}

We tabulated the data in spreadsheets for analysis. 
We used descriptive statistics to present and analyze the quantitative aspects of the extracted data and summarize the data in a comprehensible format to answer the research questions. We presented results with plots using simple numbers and sometimes means and standard deviations to help understand the results. 

To analyze the data extracted for data items adaptation problem (F7), learning problem (F8), limitations (F13) and challenges (F14), we used  
open coding~\cite{Strauss1990,Fernandez2016,Vollstedt2019} to identify the categories for each data item. In particular, we collected short descriptions of characteristic fragments from the text in the papers and identified the general concepts by labeling occurrences. This allowed us to capture the essence of what problems have been tackled by machine learning to support self-adaptation and what open problems remain. Similar to others (e.g., Prechelt et al.~\cite{Prechelt2018}), we did not have a pre-defined coding schema, but we interpreted the text in the context of the specific data items. 
To avoid bias we used the following process for coding. One researcher did a first preliminary coding. The results were then discussed among the three researchers and the codes were adjusted where needed based on consensus.

\section{Results}\label{sec:results}
We now report the results. We start with some general information and then give results grouped per research question. All data and analysis results of the study are available at the study website. 

\subsection{Demographics}
\label{subsec:primaryStudies}
By applying the search string, we retrieved \numOfIncludedPapers{} papers. After applying the inclusion/exclusion criteria on these papers, we selected \numOfSelectedPapers{} papers for data collection.\footnote{The main criteria for excluding papers were: no machine learning and no self-adaptive system (for instance papers about self-learning in the context of e-learning and self-study), no self-adaptive system (for instance papers presenting machine learning algorithms without any connection to self-adaptation), and no architecture-based adaptation (mostly papers about multi-agent systems that have no explicit distinction between managed system and managing system).} These papers were published at \numOfVenues{} venues (conferences, journals, symposia, workshops, and books).

\begin{figure}[htbp]
	\centering
	\includegraphics[width=0.95\linewidth]{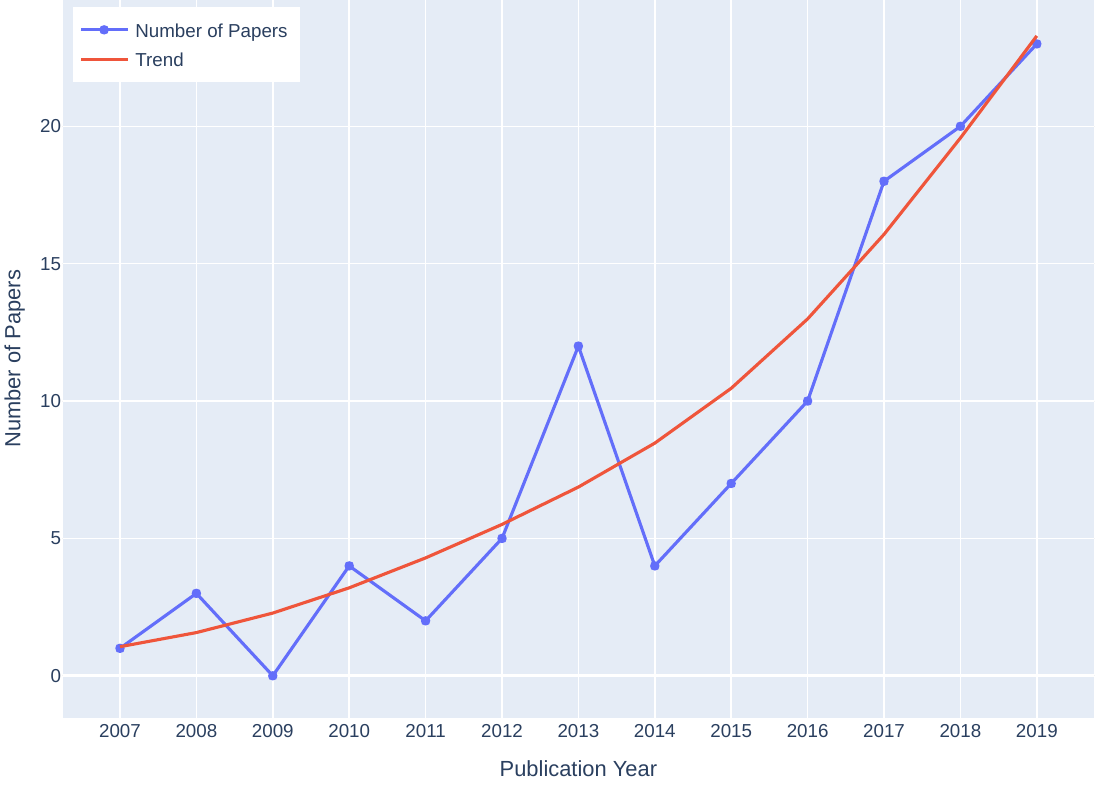}
	\caption{Distribution of selected papers over the years.}
	\label{fig:years_stats}
\end{figure}

Figure~\ref{fig:years_stats} shows the number of selected papers distributed over the years of publication. The years before 2007 are not shown since we did not find any papers before 2007. 28\% of the studies were published between 2007 and 2014 and 72\% between 2015 and 2019, which underpins the rapidly increasing research interest in the application of machine learning in self-adaptive systems.

\begin{figure}[t!]
	\centering
	\includegraphics[width=1\linewidth]{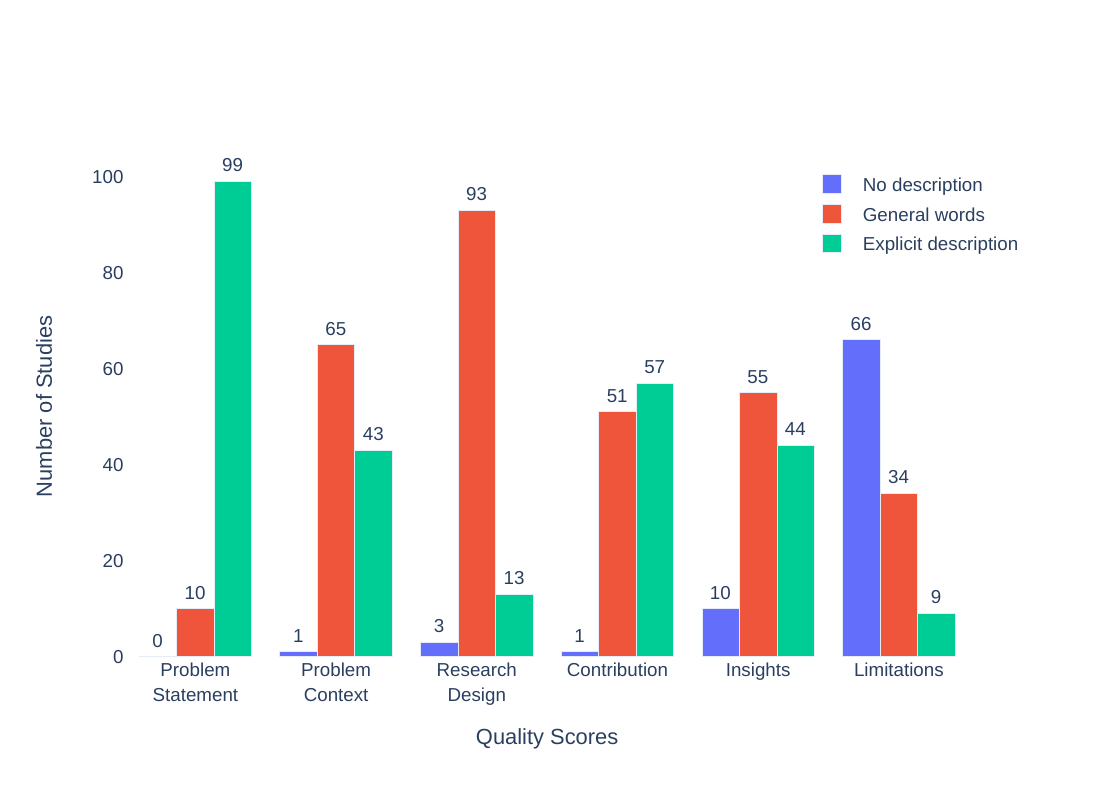}
	\caption{Quality scores of the selected papers.}
	\label{fig:quality_scores}
\end{figure}

The reporting quality of the research results for the selected papers is shown in Figure~\ref{fig:quality_scores}. The results show that most researchers provide a clear description of the problem they tackle and make clear how the problem relates to other work. Also, most of the papers clarify the contributions and discuss insights derived from the research, although not always explicitly. However, most studies ignore reporting limitations of the presented research. Similar results have been reported before in other systematic literature reviews, see for instance~\cite{Zannier06} and~\cite{MAHDAVIHEZAVEHI20171}. 

Table~\ref{tab:quality_scores} gives an overview of the different venue types with the numbers of studies for each type and the venues with the highest number of papers in each category. The table also shows the mean and standard deviation of quality scores for the different types of venues. As we can observe, the values confirm the common trend that the papers with the best quality scores are published in journals, while studies presented at workshops have lower quality scores. With a mean of the overall score of 7.7 (out of 12) and a standard deviation of 1.9, the quality of the selected papers can be regarded as reasonably good, hence providing a basis for reliable data extraction.

\begin{table}
	\centering
	\caption{Venue types with the number of studies and top venues per type, and the mean values and standard deviations of the quality scores for the different venue types (i.e., data of venues with at least three papers among the selected papers; the number or papers for each top venue is specified inside brackets).
	}
	\label{tab:quality_scores}
	\begin{tabular}{|l|c|c|c|l|}
		\hline
		\thead{Venue Type} &  \thead{\makecell{Number of \\ Studies}}  & \thead{Mean \\ Quality Score} & \thead{Standard Deviation \\ Quality Score} & \thead{\makecell{Top Venues \\ (with $\geq 3$ studies)}} \\ \hline
		Journals   & 36     & 8.6 & 1.7   & \makecell[l]{TAAS (4) \\ IEEE Access (4) \\ TSE (3) \\ Cluster Computing (3)}\\ \hline
		Conferences & 50 & 7.0 & 1.7   & \makecell[l]{ICAC (5) \\ SASO (3)}\\ \hline
		Symposia  & 13  & 7.8 & 2.4   & SEAMS (8)\\ \hline
		Books  & 4  & 8.3 & 1.0   & \\ \hline
		Workshops & 6   & 6.5 & 1.0   & FAS*W (4)\\ \hline\hline
		Overall   & 109    & 7.7 & 1.9  & \\ \hline
	\end{tabular}
	\vspace{-5pt}
	\vspace{0pt}
\end{table}

\subsection{RQ1: What problems have been tackled by machine learning in self-adaptive systems?}

To answer this research question, we analyze the data of the following data items: Adaptation problem (F7), Learning problem (F8), Application domain (F12).\footnote{During the coding process, we used the following terminology. We used the term \textit{quality} to refer to non-functional requirements that describe how a system should perform its functions, such as its performance and reliability. We used the term \textit{resource} to refer to the means or supplies a software system uses to realize its functions, such as memory and bandwidth. We used the term \textit{cost} to refer to monetary aspects, i.e., the price one has to pay to use or operate a system.}  
\vspace{10pt}

\noindent
\textbf{Adaptation problem.} In a concrete setting, self-adaptation is applied to solve a particular adaptation problem. This adaptation problem refers to the concerns that the managing system is dealing with. Previous research has shown that these concerns relate to quality properties of, and resources used by the system, see e.g.,~\cite{Villegas2011,weyns2013claims}. Table~\ref{tab:adaptation problems} lists the adaptation problems we have identified from the papers, illustrated with examples. The last column shows the number of papers for each type.

\setlength\cellspacetoplimit{10pt}
\setlength\cellspacebottomlimit{6pt}
\begin{table*}[h!]
	\centering
	\caption{Adaptation problems illustrated with examples, with the frequencies of papers (right column).}\vspace{-5pt}
	\label{tab:adaptation problems}
	\renewcommand{\arraystretch}{1.2}
	\begin{tabu}{|X[1,m]X[2,m]|X[0.25,c,m]|} 
		\hline
		\textbf{Adaptation Problem}     &
		\textbf{Brief Description with Example}     &
		\textbf{\#} \\ 
		\hline
		\multirow{1}{*}{\shortstack[l]{Improve qualities}} &
		\Tstrut 
		This adaptation problem is about improving (i.e., optimizing, maintaining, etc.) quality properties of the system. An example is keeping the response time low and the reliability high under changing workload and the occurrence of unexpected events~\cite{Elkhodary2010}.
		\Bstrut
		& 44
		\\  \tabucline[1pt]{-}
		
		\multirow{1}{*}{\shortstack[l]{Balance qualities \\ with resources}} & 
		\Tstrut
		This adaptation problem is about keeping a balance between improving quality properties of the system and  resources (i.e., CPU, energy, etc.) required to achieve that improvement. For example,\cite{Maggio2012} aims at keeping the latency within a specified range while minimizing the computational resource required to achieve that. 
		\Bstrut
		& 37
		\\  \tabucline[1pt]{-} 
		
		\multirow{1}{*}{\shortstack[l]{Balance qualities \\ with cost}} & 
		\Tstrut
		This adaptation problem is about keeping a balance between improving the system's quality properties and the cost (i.e., operational, financial, etc.) required to achieve that. For example, \cite{Calinescu11} aims at keeping the failure rate of a service-based workflow below a specified threshold while minimizing the cost for using the services.
		\Bstrut
		& 15
		\\  \tabucline[1pt]{-} 
		
		\multirow{1}{*}{\shortstack[l]{Improve resource allocation}} & 
		\Tstrut
		This adaptation problem is about improving (i.e., optimizing, managing, etc.) the system's resource consumption. An example is managing the system's CPU and memory usage under uncertain workloads of the system~\cite{Jamshidi16}.
		\Bstrut
		& 10
		\\  \tabucline[1pt]{-} 
		
		\multirow{1}{*}{\shortstack[l]{Protect against cyber threats}} & 
		\Tstrut
		This adaptation problem is about automating the cyber defense of a system by detecting and managing threats (i.e., intrusion, anomalies, etc.). For instance, \cite{Maimo2018} considers cyber defence in  fifth-generation mobile systems by detecting and dealing with system intrusions.
		\Bstrut
		& 3
		\\  \hline
		
	\end{tabu}
	\vspace{8pt}
\end{table*}

\noindent 
\textbf{Learning problem.} A learning problem refers to a concrete problem that needs to be solved by machine learning in support to realize self-adaptation. Table~\ref{tab:learning problems} shows the six types of learning problems that we have identified from the papers. Each type is illustrated with examples. The last column shows the number of papers for each type of learning problem.

\setlength\cellspacetoplimit{10pt}
\setlength\cellspacebottomlimit{6pt}
\begin{table*}[h!]
	\centering
	\caption{Learning problems illustrated with examples, with the frequencies of papers (right column).}\vspace{-5pt}
	\label{tab:learning problems}
	\renewcommand{\arraystretch}{1.2}
	\begin{tabu}{|X[1,m]X[2,m]|X[0.25,c,m]|} 
		\hline
		\textbf{Learning Problem}     &
		\textbf{Brief Description with Example}     &
		\textbf{\#} \\ 
		\hline
		\multirow{1}{*}{\shortstack[l]{Update/Change \\ adaptation rules/policies}} & 
		\Tstrut 
		This learning problem is about updating or changing adaptation rules or policies to support a managing system when dealing with changing operating conditions.   For example, in~\cite{Gu2012}, the managing system is supported by a  reinforcement learner that dynamically updates adaptation policies to deal with changing workload.  
		\Bstrut
		& 36
		\\  \tabucline[1pt]{-}
		
		\multirow{1}{*}{\shortstack[l]{Predict/Analyze \\ resource usage}} & 
		\Tstrut
		This learning problem is about  predicting or analyzing  resources that are used by the managed system that affect the decision-making of the managing system. Examples are learners that predict the energy consumption of batteries~\cite{LiuTeng2018}, storage~\cite{El19}, and CPU usage~\cite{El19}. 
		\Bstrut
		& 23
		\\  \tabucline[1pt]{-} 
		
		\multirow{1}{*}{\shortstack[l]{Keep runtime models \\ up-to-date}} & 
		\Tstrut
		This learning problem is about supporting a managing system with keeping runtime models up-to-date. Examples are a model of the environment~\cite{Sykes2013}, a performance model~\cite{Jamshidi2018}, and a reliability model~\cite{Calinescu2014}.
		\Bstrut
		& 18
		\\  \tabucline[1pt]{-} 
		
		\multirow{1}{*}{\shortstack[l]{Reduce \\ large adaptation space}} & 
		\Tstrut
		This learning problem is about supporting a managing system with reducing a large number of adaptation options (large adaptation space) such that the system can make more efficient decisions. For instance,~\cite{Sommer2015} and \cite{Quin2019} use learners to predict quality properties of adaptation options to select options, speeding up analysis.
		\Bstrut
		& 16
		\\  \tabucline[1pt]{-} 
		
		\multirow{1}{*}{\shortstack[l]{Detect/Predict anomalies}} & 
		\Tstrut
		This learning problem is about detecting or predicting anomalies in the behavior of the system or its environment that are relevant for adaptation. For example, in~\cite{Krupitzer2017} a learner detects abnormal flow of traffic in a traffic management system and in~\cite{Papamartzivanos2019} a learner identifies cyber threats in a communication network. 
		\Bstrut
		& 12
		\\  \tabucline[1pt]{-} 
		
		\multirow{1}{*}{\shortstack[l]{Collect \\ unavailable prior knowledge}} & 
		\Tstrut
		This learning problem is about collecting initially unknown runtime knowledge to support adaptation. For instance, in~\cite{Ghahremani2018} a learner builds a performance model to support the managing system with computing the utility of 
		different configurations;  in~\cite{tesauro2007use} a learner  identifies management policies without any prior knowledge. 
		\Bstrut
		& 4
		\\  \hline
		
	\end{tabu}
	\vspace{8pt}
\end{table*}

\begin{figure}[htbp]
	\centering
	\includegraphics[width=0.9\linewidth]{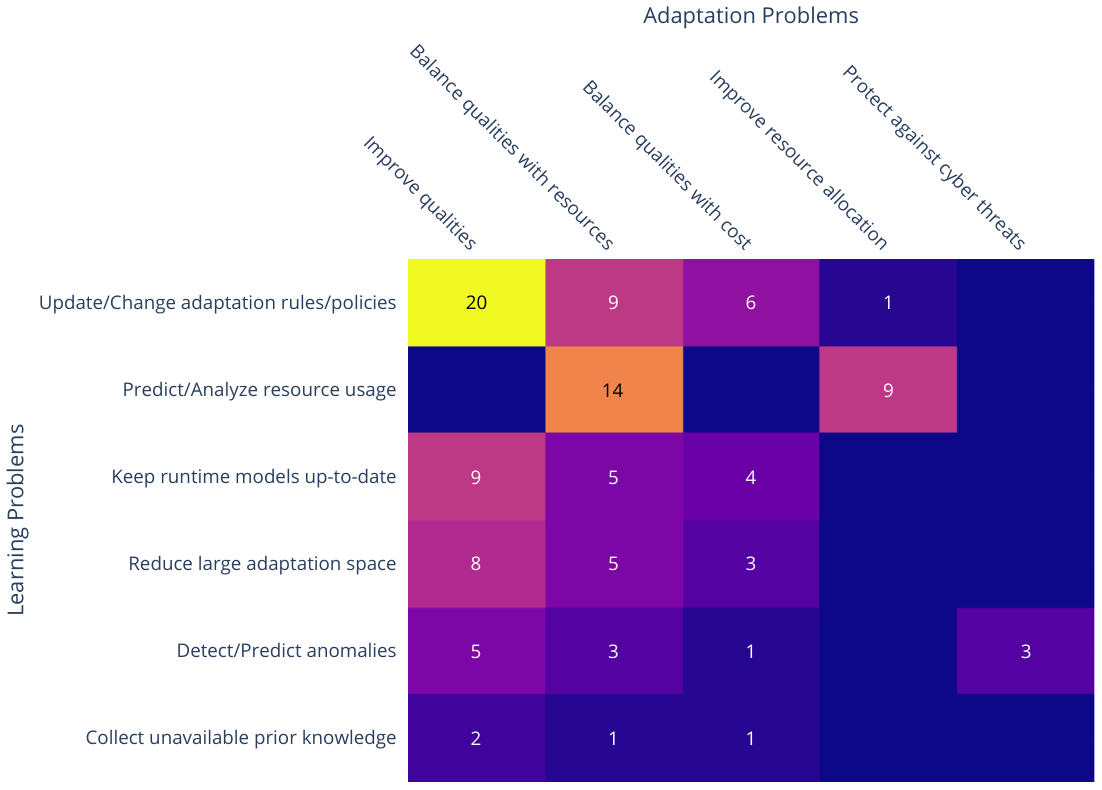}
	\caption{Learning problems versus adaptation problems.}
	\label{fig:adaptation vs learning problems}
\end{figure}

\noindent 
\textbf{Adaptation problems versus Learning problems.} 
We can now map adaptation problems to learning problems. This mapping allows us to identify whether particular types of adaptation problems delegate particular sub-problems to a machine learner. Figure~\ref{fig:adaptation vs learning problems} shows the mapping.

A few observations jump out. First, self-adaptive systems that aim at improving qualities of the managed system primarily use learning to solve the problem of updating and changing adaptation rules and policies. Second, self-adaptive systems that aim at protecting against cyber threats exploit learning only to detect or predict anomalies. Third, self-adaptive systems that aim at balancing qualities with resources used by the system exploit learners to solve all types of learning problems.

\noindent
\textbf{Application domain.} Table~\ref{tab:application_domains} shows the application domains where machine learning has been applied and evaluated in support of self-adaptation. The results show that learning has been applied in a wide variety of application domains, yet  over 60\% of the papers have studied and validated their work in three domains: cloud, client-server systems, and cyber-physical systems. 

\begin{table}[htbp]
	\centering
	\caption{Application domains of the collected papers.}
	\label{tab:application_domains}
	\vspace{-5pt}
	\begin{tabular}{|c|c|}
		\hline
		\textbf{Application domain}     & \textbf{Number} \\ \hline
		Cloud                 &   	33 \\ \hline
		Client-server system  &	    18 \\ \hline
		Cyber-physical system &  	16 \\ \hline
		Internet-of-things    &	     9 \\ \hline
		Service-based system  &      8 \\ \hline
		Robotics	          &      7 \\ \hline
		Network management	  &      6 \\ \hline    	
		Business process management
		&  	 4 \\ \hline
		Remote data mirroring &      3 \\ \hline
		Traffic management	  &      3 \\ \hline
		Stream processing     
		&      2 \\ \hline
		Grid computing	      &      1 \\ \hline	     
		Medical simulation &      1 \\ \hline
		No specific domain  &      3 \\ \hline
	\end{tabular}
	\vspace{5pt}
\end{table}

\vspace{10pt}

\begin{figure}[htbp]
	\centering
	\includegraphics[width=0.9\linewidth]{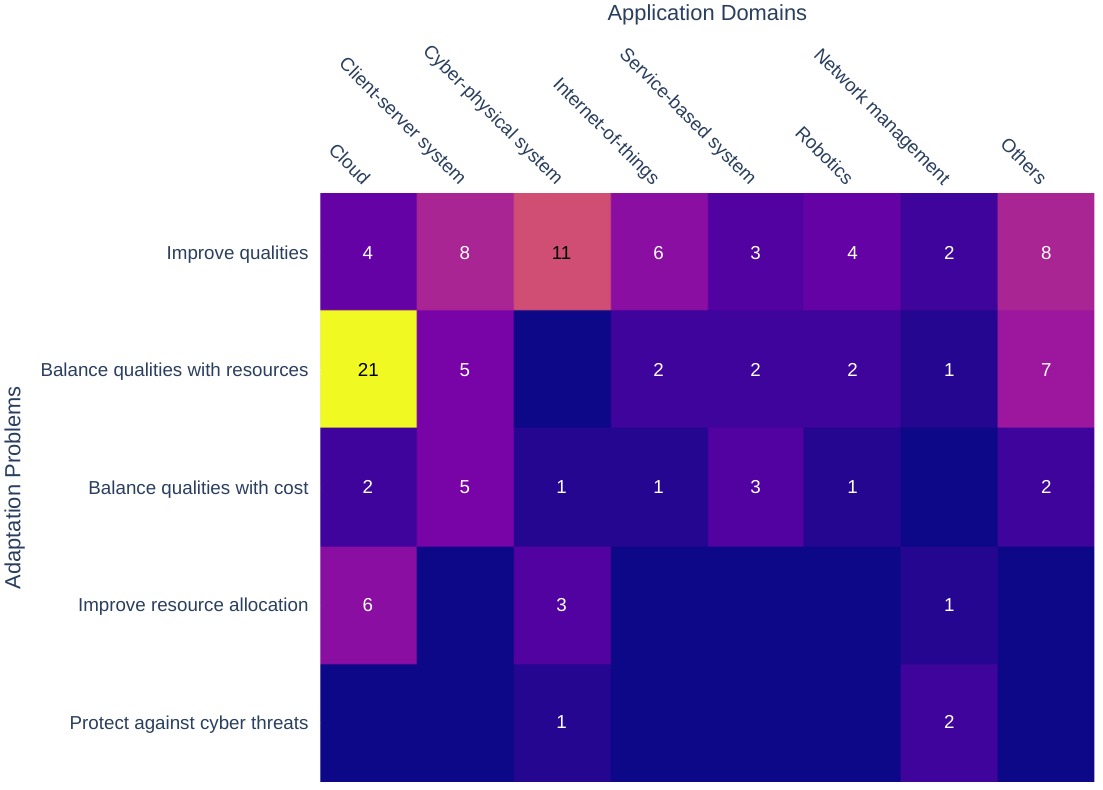}
	\caption{Adaptation problems versus Application domains.}
	\label{fig:adaptation problems vs application domains}
\end{figure}

\noindent
\textbf{Application domains versus Adaptation problems.}
Figure~\ref{fig:adaptation problems vs application domains} shows the adaptation problems solved in different application domains. Improving qualities is the main adaptation problem considered in all domains, except cloud. This can be expected as managing resources is vital in cloud applications, so balancing qualities with resources is the dominant adaptation problem in the cloud domain. Surprisingly, only a small fraction of the studies in the domains of cyber-physical systems and the internet-of-things consider resource management as part of the adaptation problem. Cost as an explicit factor in self-adaptation is mainly considered in client-server systems and service-based systems (in terms of the price to pay for using services). 

\vspace{10pt}

\begin{figure}[htbp]
	\centering
	\includegraphics[width=0.9\linewidth]{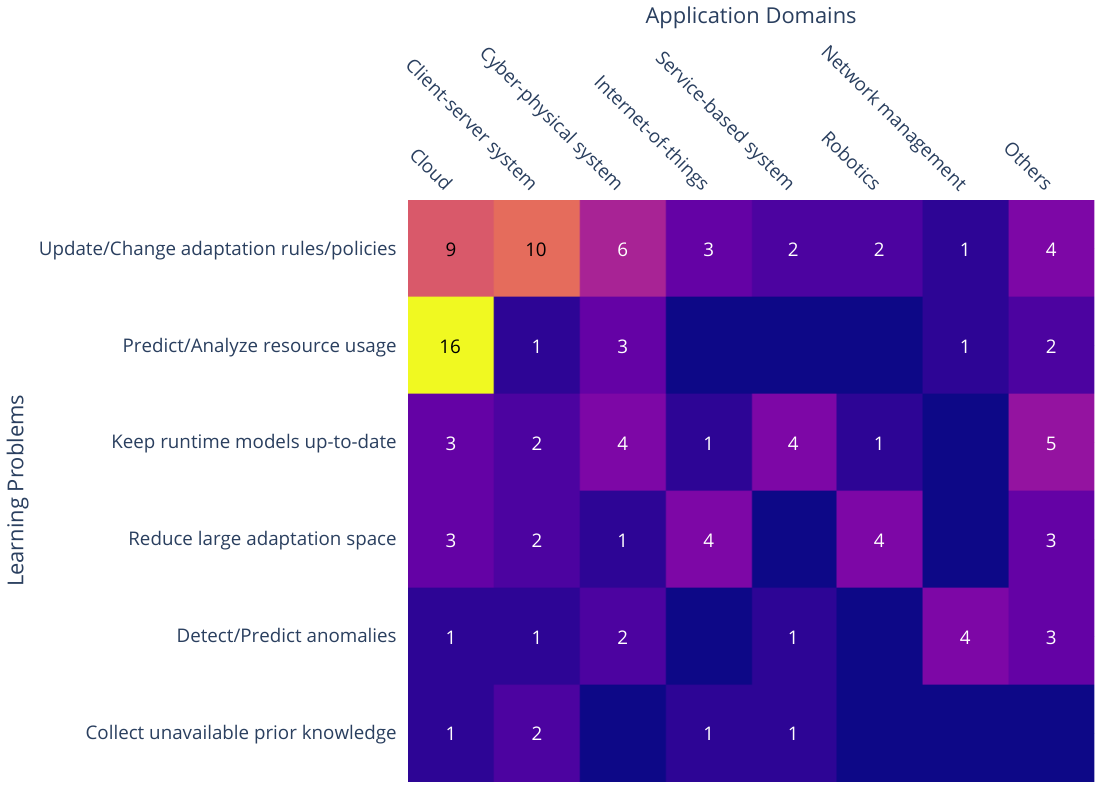}
	\caption{Learning problems versus Application domains.}
	\label{fig:learning problems vs application domains}
\end{figure}

\noindent
\textbf{Application domains versus Learning problems.}
It is also interesting to take a look at the mapping of solved learning problems in different application domains, as shown in Figure~\ref{fig:learning problems vs application domains}. One key observation is that learning for updating and changing adaptation rules and policies is used in all domains, with cloud and client-server systems as main domains. 
The latter resonates with the broad use of rule-based and policy-based techniques in the two domains. Learning to predict and analyze resource usage on the other hand is primarily used in the cloud domain only. Learning to keep runtime models up to date as well as reducing large adaptation spaces are also broadly used across domains (except in network management). Finally, detecting and predicting anomalies is primarily applied in network management, but sporadically also in a variety of other domains.

\vspace{0.3cm}
\begin{tcolorbox}[breakable]
	\textbf{Answer to RQ1 - What problems have been tackled by machine learning in self-adaptive systems.} 
	We identified five types of adaptation problems where learning is applied: \emph{improve qualities}, \emph{balance qualities with resources}, \emph{balance qualities with cost}, \emph{improve resource allocation}, and \emph{protect against cyber threats}. 
	We identified six types of learning problems: \emph{update/change adaptation rules/policies}, \emph{predict/analyze resource usage}, \emph{keep runtime models up-to-date}, \emph{reduce large adaptation space}, \emph{detect/predict anomalies}, and \emph{collect unavailable prior knowledge}. The dominant case where learning is used to solve adaptation problems is updating and changing adaptation rules and policies to support improving qualities of the system. Learning to support self-adaptation has been applied in a variety of applications, with cloud, client-server system, and cyber-physical system as main domains. 
\end{tcolorbox}

\vspace{3pt}\noindent 
\subsection{RQ2: What are the key engineering aspects considered when applying learning in self-adaptation?}

To answer this research question, we use the data items: MAPE stage(s) supported by learning (F9), Dimensions of learning methods (F10), and Learning methods used to support self-adaptation (F11).

\vspace{3pt}\noindent 
\textbf{MAPE functions supported by learning.} 

\begin{figure}[htbp]
	\centering
	\vspace{5pt}
	\includegraphics[width=0.7\linewidth]{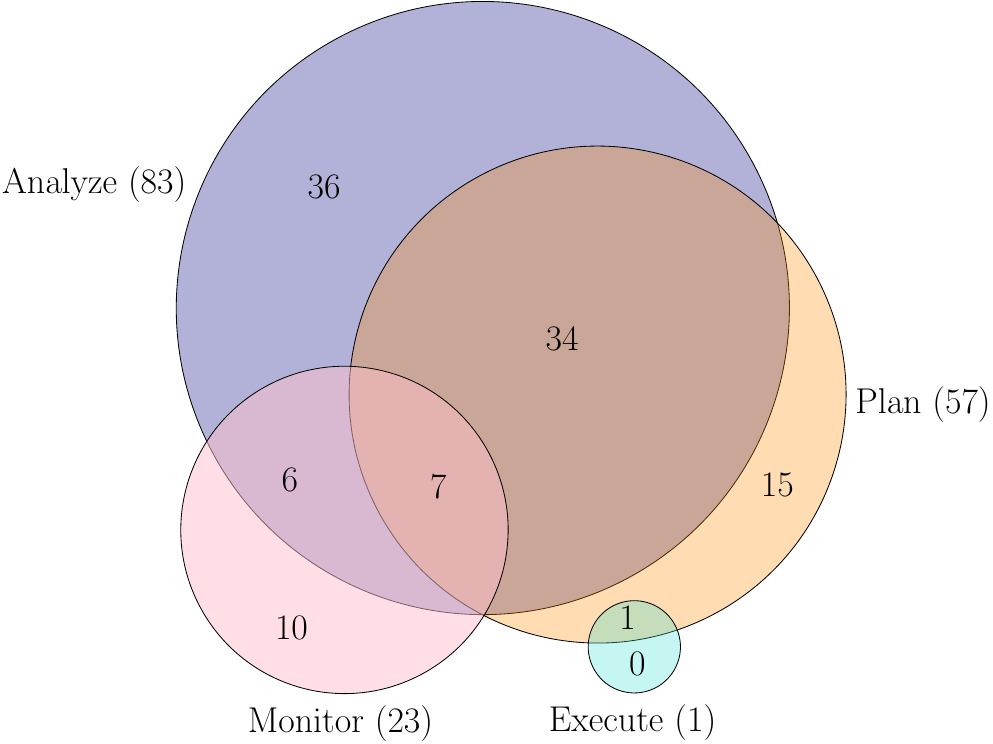}
	\caption{Distribution of learning methods supporting the MAPE functions.}  
	\label{fig:ml_mape_relation}
\end{figure}

Figure~\ref{fig:ml_mape_relation} shows the distribution of learning methods used to support MAPE functions. The diagram shows that learning is dominantly used to support the analysis part of the decision-making process in the feedback loop (in 83 studies). Thirty-six of these studies apply learning in support of analysis only. A typical example is~\cite{Quin2019}, where machine learning is used to reduce large adaptation spaces such that only the relevant options need to be analyzed. 
Besides analysis, learning is often used to support planning of the feedback loop (57 studies). Fifteen of these studies apply learning in support of planning only.
For example, \cite{Pandey2017} adopted an instance-based learning method to implement a hybrid planning approach that selects an optimal planning strategy among possible strategies. 
Learning has also been used to support monitoring (23 studies), in particular to update knowledge models. 
Ten of these studies apply learning in support to monitoring only.
For example, in~\cite{Kramer2012}, monitoring data is pre-processed using a light-weight classifier in order to learn optimization rules. We only identified one paper~\cite{Papamartzivanos2019} where machine learning was used to support the execution function of the feedback loop (in combination with planning). In this paper, the authors used a classifier to choose the actuator that the system should use to adapt. The diagram shows that in a substantial number of papers (34 in total), learning supports both analysis and planning. An example is described in~\cite{Frommgen2015}, where machine learning is applied to generate Event-Condition-Action rules that are evaluated and subsequently used to make adaptation decisions. A smaller fraction of the papers (six in total) combine learning to support monitoring and analysis. Finally, a small number of papers (seven in total) apply learning that spans monitoring, analysis, and planning. As an example, \cite{Bierzynski19} proposed a proactive learner that supports monitoring, analysis, and planning by collecting context data, extracting required data for updating the learning model, and preparing a reasoning module for the decision-making process.

\begin{figure}[htbp]
	\centering
	\includegraphics[width=0.85\linewidth]{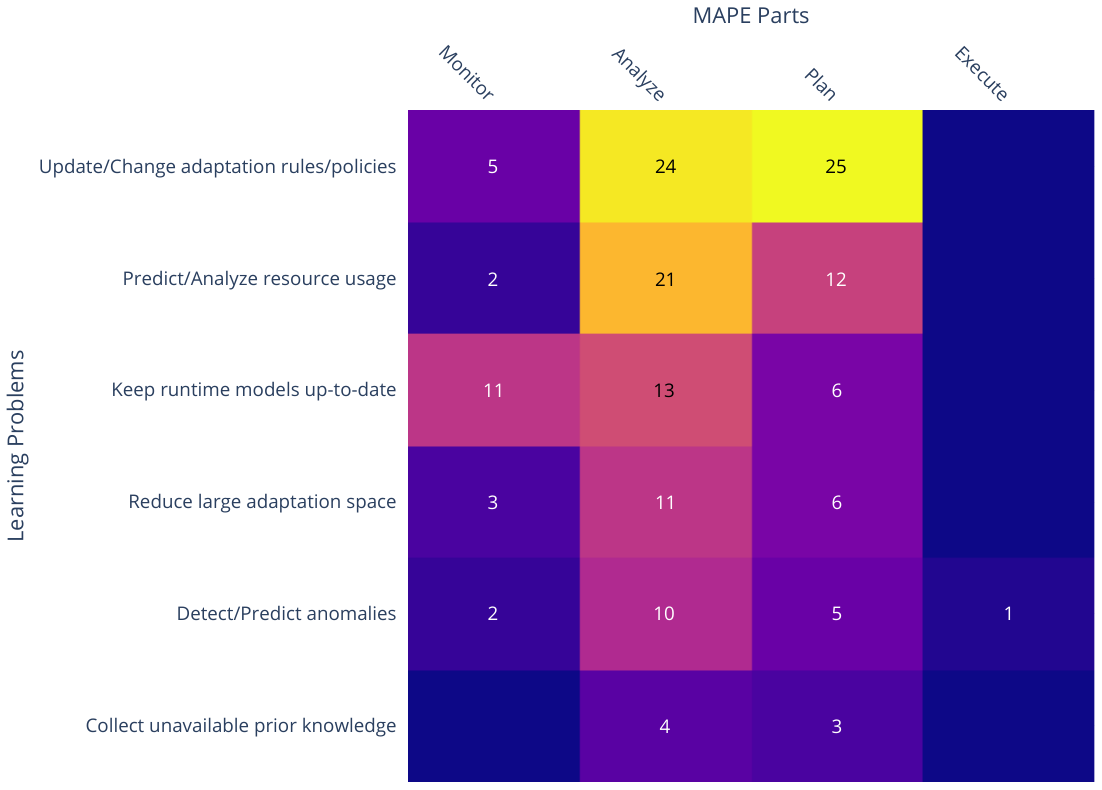}
	\caption{Learning problems vs. MAPE functions.}
	\label{fig:learning problems vs. MAPE}
\end{figure}

\vspace{3pt}\noindent 
\textbf{Learning problems versus MAPE functions.} %
Figure~\ref{fig:learning problems vs. MAPE} maps the learning problems to MAPE functions. The heat-map shows that all types of learning problems are tackled in support of monitoring, analysis, and planning. As can be expected, the dominant case is learning used for updating and changing adaptation rules and policies to support analysis and planning, followed by predicting and analyzing resource usage. Another observation is the importance of learning used to keep runtime models up-to-date in support for monitoring and analysis. 

\begin{figure}[htbp]
	\centering
	\includegraphics[width=1\linewidth]{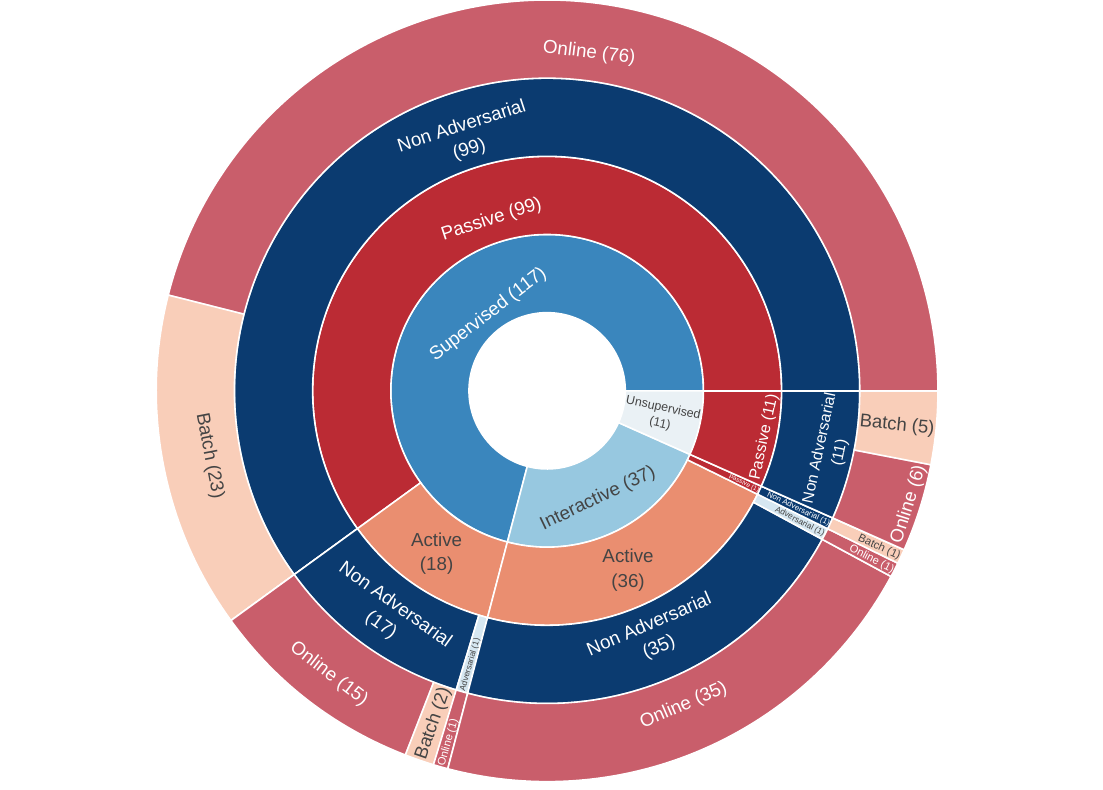}
	\caption{Distribution of learning dimensions used in machine learning for self-adaptive systems}
	\label{fig:learning_paradigms}
\end{figure}

\vspace{3pt}\noindent 
\textbf{Learning dimensions.} Figure~\ref{fig:learning_paradigms} gives an overview of the learning dimensions that have been applied in the papers. Each layer of the sunburst diagram shows the options for one of the learning dimensions, where the same value of a dimension is represented by the same color. 
The numbers in the diagram are based on the  number of learning tasks that are solved to support self-adaptation. For instance, six learning tasks have been solved using one or more learning methods that apply online, non adversarial, passive, unsupervised learning. In total 165 learning tasks have been solved in the 109 papers using a variety of learning methods. We zoom in on the concrete learning tasks solved by concrete learning methods below. 
The results show that a wide variety of combinations of dimensions are applied. The most popular learning methods used in self-adaptation apply supervised, passive, non adversarial, and online learning. In terms of learning type, we observe that supervised learning dominates (71\% of the learning tasks). On the other hand, only a small number of papers apply unsupervised learning (7\% of the learning tasks), which is surprising for self-adaptive systems that aim at automation and dealing with uncertainties that may not have been anticipated\,\cite{9196226}. Example papers that applied unsupervised learning are \cite{Duarte2018} and \cite{vzapvcevic13}, in particular clustering-based learning techniques. 
Another observation is that we only encountered two papers that use adversarial learning, namely \cite{Khan2017} and \cite{lee2019nash} that applied a game-theoretical learning approach. 

\begin{figure}[htbp]
	\centering
	\includegraphics[width=0.88\linewidth]{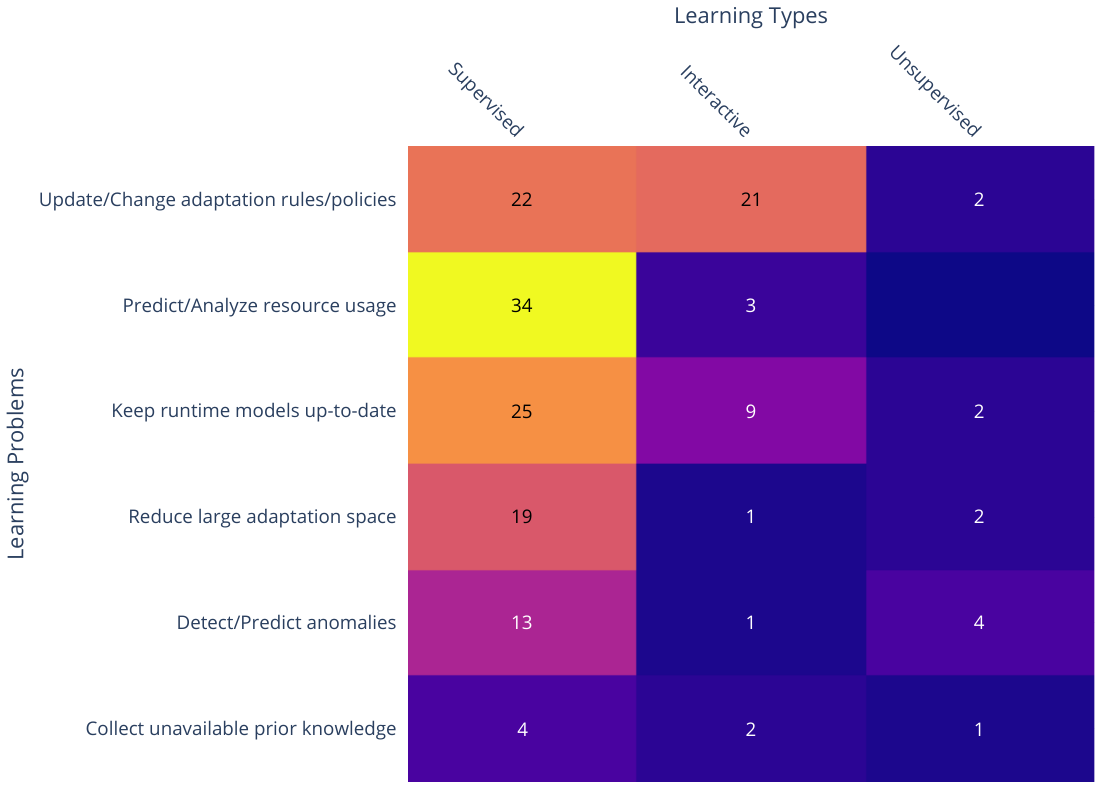}
	\caption{Learning problems vs. Learning types.}
	\label{fig:learning_vs_data}
\end{figure}

\vspace{3pt}\noindent 
\textbf{Learning problems versus learning types.} 
Figure~\ref{fig:learning_vs_data} shows the mapping of learning problems to learning types. 
The results demonstrate that supervised learning is frequently used for all types of learning problems, but mostly to predict and analyze resource usage. Interactive learning is also used for all types of learning problems, yet updating and changing adaptation rules and policies together with keeping runtime models up-to-date make up 80\% of these problems. Unsupervised learning is most frequently used for detecting and predicting anomalies, but nevertheless supervised is still used three times more to tackle this learning problem.   

\begin{figure}[htbp]
	\centering
	\includegraphics[width=1.05\linewidth]{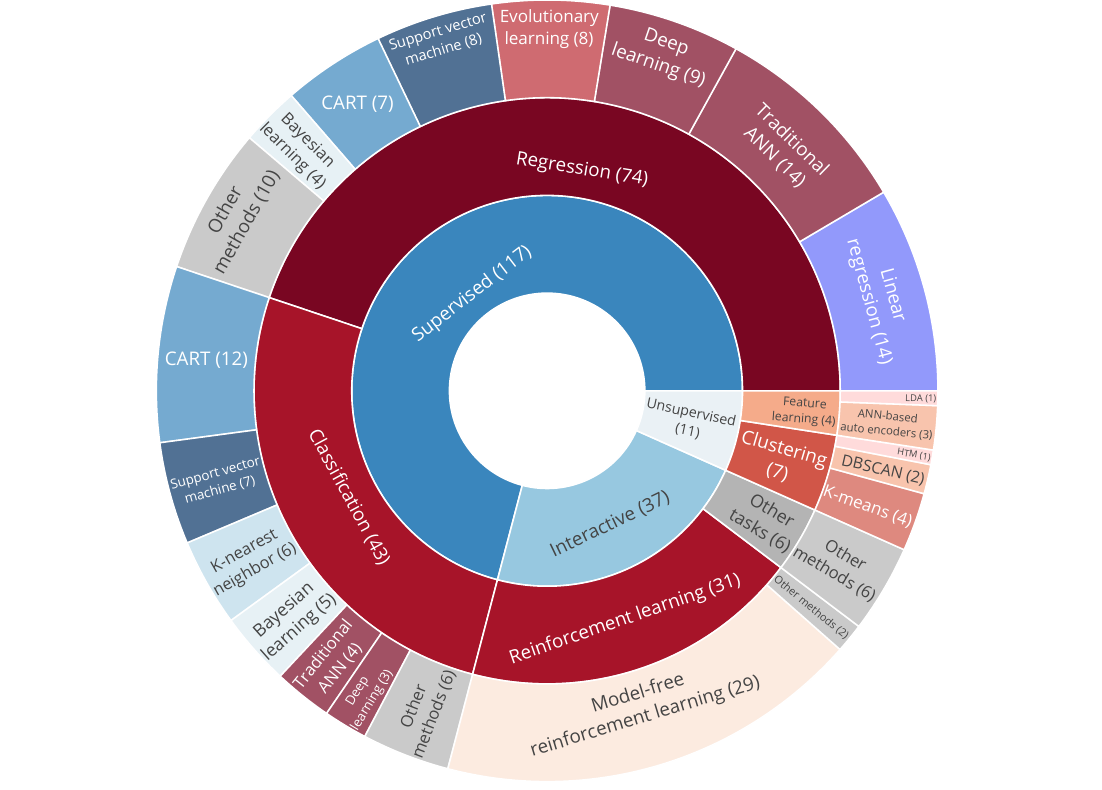}
	\caption{Distribution of learning types, tasks, and methods used in self-adaptive systems. ANN refers to Artificial Neural Network, HTM to Hierarchical Temporal Memory and LDA to Latent Dirichlet Allocation.
	}
	\label{fig:ml_methods}
\end{figure}

\vspace{3pt}\noindent 
\textbf{Learning methods used to support self-adaptation.} 
Figure~\ref{fig:ml_methods} shows an overview of concrete learning methods used in self-adaptation. Note that multiple methods may be used in a single study. Each layer of the sunburst diagram shows a different level of abstraction of the learning methods. The inner layer groups methods based on learning type, i.e., the dimension ``unsupervised vs. supervised vs. interactive'' (we further elaborate on dimensions of the learning methods below). The layer in the middle groups learning methods based on common tasks of machine learning methods, i.e., classification, regression, reinforcement learning, clustering, and feature learning. This grouping is based on~\cite{shalev2014understanding, bishop2006pattern}. Note that one method can be used for multiple tasks, for instance support vector machines and deep learning have been used for both regression and classification. Finally, the outer layer shows concrete learning methods that were used in support of self-adaptation together with their frequencies (numbers between brackets). 
Areas marked with ``Other tasks/methods'' group other options. For instance out of the 37 papers with interactive learning methods, 31 use reinforcement learning to solve a learning problem; the other six studies use other methods, such as hidden semi-Markov models and partially observable Markov decision process. The diagram shows that the dominating learning method used in support of self-adaptation is model-free reinforcement learning (29 papers). An example is \cite{Arabnejad16} that utilized fuzzy Q-learning to reason about new rules from the data collected at runtime. 
 Other popular learning methods used in self-adaptation are support vector machines (15 times used; eight for regression and seven for classification), and traditional artificial neural networks and linear regression (both 14 times used). For instance, \cite{Elgendi2019} exploited a support vector machine to detect network attacks in cyber-physical systems, \cite{Chen17} used an artificial neural network to predict qualities of services such as response time and throughput of the system, and \cite{Sensi16} applied linear regression to predict performance and power consumption.

\begin{figure}[htbp]
	\centering
	\includegraphics[width=0.9\linewidth]{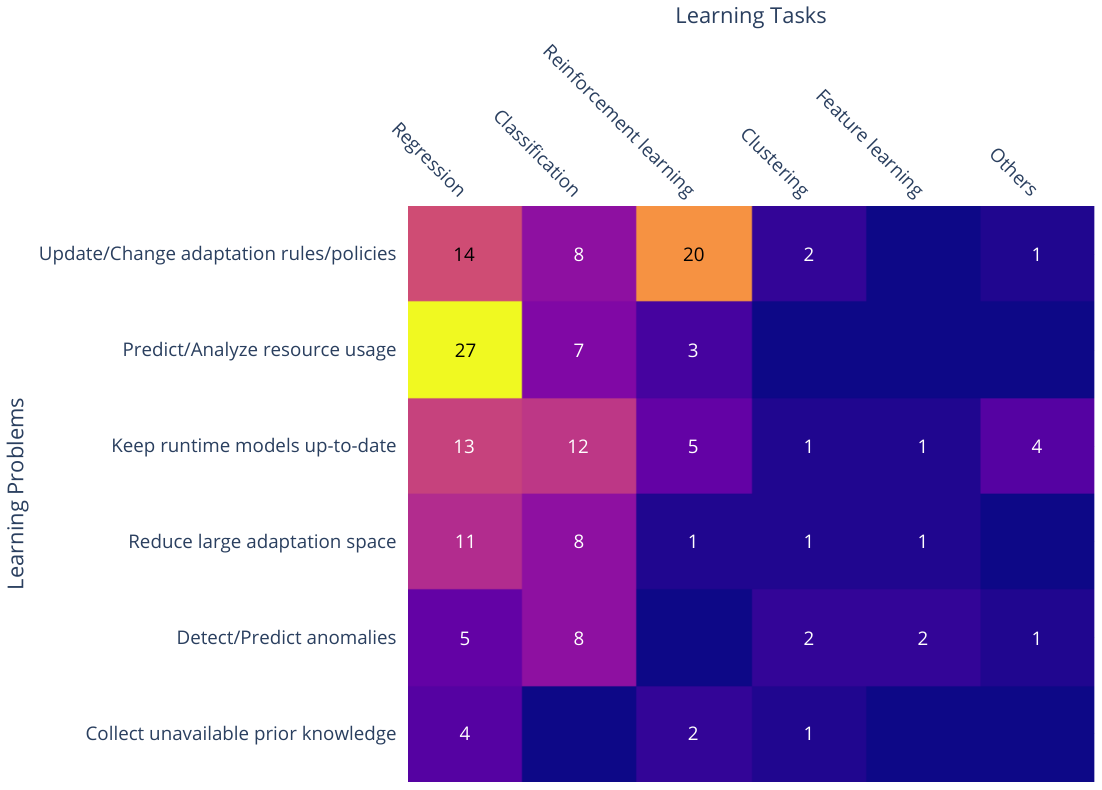}
	\caption{Learning problems vs. Learning tasks.}
	\label{fig:pvsalg}
\end{figure}

\vspace{3pt}\noindent 
\textbf{Learning problems versus learning tasks.} 
We also looked at the mapping between the learning problems in support of self-adaptation and the different types of learning tasks that need to be solved by the learners. Figure~\ref{fig:pvsalg} shows an overview of this mapping. Regression is frequently used to solve all types of learning problems, but mostly to predict and analyze resource usages (36\% of the problems solved with regression). Classification is also broadly used, with keeping runtime models up-to-date as the main learning problem (28\% of the problems solved with a classifier). Updating and changing adaptation rules and policies is the primary learning problem solved by reinforcement learning (65\% of the problems solved by a reinforcement learner). 

\begin{figure}[htbp]
	\centering
	\includegraphics[width=0.9\linewidth]{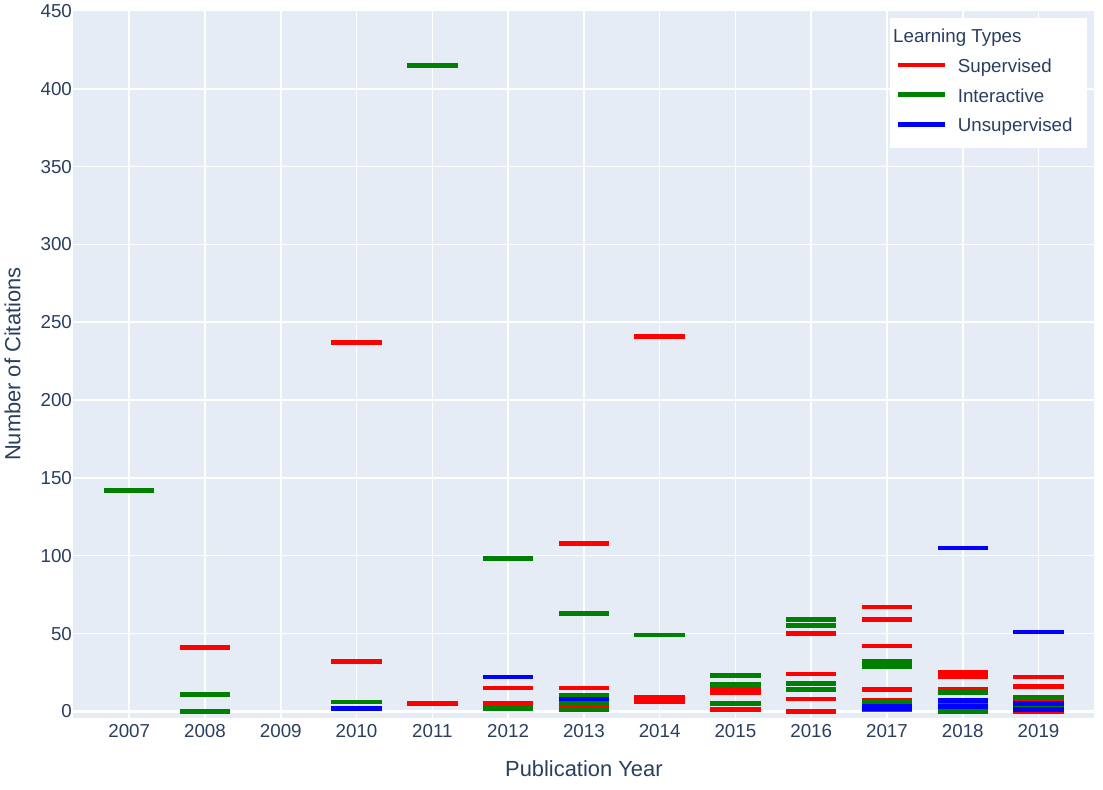}
	\caption{ Distribution of learning types through years and number of citations (Google Scholar 2/2022). Each study is represented by a tiny horizontal bar as indicated in the key. Studies of the same learning type with citation counts close to each other form thicker bars.}
	\label{fig:years_citations}
\end{figure}

\vspace{3pt}\noindent 
\textbf{Distribution of learning tasks over time and number of citations.} 
To conclude, we look at the distribution of learning types over the years and the impact of the papers based on the number of citations they generated. For the latter we used the citations of Google Scholar February 2021.\footnote{We used Google Scholar as it is widely used, but we acknowledge its limitations, such as the inclusion of self-citations.} Figure~\ref{fig:years_citations} plots the results. We observe that only seven papers have generated more than 100 citations; four that used supervised learning~\cite{Maimo2018, Esfahani2013, Elkhodary2010, Zuo2014}, two that used interactive learning~\cite{tesauro2007use, Calinescu11}, and one paper that used unsupervised learning~\cite{Maimo2018}. Overall, none of the three learning types seem to have generated clearly more impact (normalized number of citations\footnote{The number of citations for each paper has been normalized by past years since it was published, i.e., normalized number of citations of paper = Google Scholar citation of the paper in 2021/(2021 - publication year of the paper).} 
2007-2019 for supervised learning avg 4.8, std 7.0; interactive learning avg 4.4, std 7.4; and unsupervised learning avg 6.5, std 11.4.). The plot shows that supervised and interactive learning have been used frequently over the full time span from 2007 till 2019. Yet, since 2016, we note an increase in the use of supervised learning and a decrease in interactive learning. Remarkably, after some small  attention on the use of unsupervised learning in the period 2010 to 2013 (three studies), we observe an increase for this type of learning in the last three years, from 2017 to 2019 (eight studies).

\vspace{0.3cm}
\begin{tcolorbox}[breakable]
	\textbf{Answer to RQ2 - What are the key engineering aspects considered when applying learning in self-adaptation?} Machine learning is primarily used to support analysis and planning in self-adaptive systems.  
	The majority of the papers use supervised or interactive learning; these learners 
	typically exploit results of runtime analysis and observed effects of applied adaptations to learn. 
	The most frequent problem of self-adaptation delegated to learning is updating and changing adaptation rules and policies (primarily solved using regression and reinforcement learning). Other important learning problems are predicting and analyzing resource usage (primarily solved using regression) and keeping runtime models up-to-date (primarily solved using regression and classification). The most popular learning method that is applied in self-adaptive systems is model-free reinforcement learning used for updating and changing adaptation rules and policies. Adversarial learning on the other hand is understudied, while this approach has a huge potential to deal with security concerns in self-adaptation.   We observe that supervised and interactive learning have been used frequently over the years. Unsupervised learning on the other hand  has only been used in a limited number of papers. Yet, this approach supports detecting novelty in data without any labeling, which can play a key role in managing complex types of uncertainty.  None of the three types of learning has clearly generated more impact over the years. 
\end{tcolorbox}

\subsection{RQ3: What are open challenges for using machine learning in self-adaptive systems?}

To answer this research question, we analyze data items: Limitations (F13) and Challenges (F14). 

\vspace{3pt}\noindent 
\textbf{Limitations.} Table~\ref{tab:limitations} lists the limitations reported in the papers. As illustrated in Figure~\ref{fig:quality_scores}, only 
a limited number of papers 
reported limitations of the applied learning methods. 
The results show that a variety of limitations of the learning methods have been reported. The most frequently reported limitation is limited scalability of the proposed learning approach. Other reported limitations relate to impact on qualities, in particular performance and reusability, the scope in terms of uncertainty and guarantees that can be provided, and the need for expertise of humans to tune parameters.

\begin{table*}[h!]
	\centering
	\caption{Reported limitations of the learning methods applied in the papers.}    
	\label{tab:limitations}
	\vspace{-5pt}
	\renewcommand{\arraystretch}{1.2}
	\begin{tabu}{|X[0.75,m]X[1.5,m]|X[0.15,c,m]|X[0.6,c,m]|} 
		\hline
		\multirow{2}{*}{\shortstack[l]{Scalability}} & Learning approach is not scalable & 6 & \cite{Calinescu2014, Papamartzivanos2019, Chen17, Calinescu2017, tesauro2007use, tang2018io}
		\\ \cline{2-4}
		& No test of scalability due to data sensitivity & 1 & \cite{Salfner2010} \\ \tabucline[1pt]{-}
		\multirow{3}{*}{\shortstack[l]{Performance}} & High computation time & 3 & \cite{Krupitzer2018, dhrgam2018, tesauro2007use}
		\\ \cline{2-4}
		& High computational load for feedback loop  & 1 & \cite{Gerostathopoulos16} \\
		\cline{2-4}
		& Slow convergence & 1 & \cite{Maggio2012}  \\
		\tabucline[1pt]{-}
		\multirow{2}{*}{\shortstack[l]{Reusability}} & Solution is domain specific & 2 & \cite{Ferroni2017, Gerostathopoulos16} 
		\\ \cline{2-4}
		& Applicable for optimization problems only & 1 &  \cite{LiuYang2018}
		\\
		\tabucline[1pt]{-}
		\multirow{2}{*}{\shortstack[l]{Uncertainty}} & Cannot handle new situations & 2 & \cite{Feng2015, Sykes2013}
		\\ \cline{2-4}
		& Cannot detect sudden changes  & 1 & \cite{tang2018io}  \\
		\tabucline[1pt]{-}
		
		\multirow{2}{*}{\shortstack[l]{Guarantees}} & Optimization without satisfying all SLAs  & 1 & \cite{Qin12} 
		\\ \cline{2-4}
		& Might be trapped in a sub-optimal solution  & 1 & \cite{qian2015rationalism} \\
		\tabucline[1pt]{-}
		Design & Need parameter tuning & 4 & \cite{Maggio2012, Feng2015, Maimo2018, Jamshidi2016} \\ \hline
	\end{tabu}
	\vspace{8pt}
\end{table*}

\vspace{3pt}\noindent 
\textbf{Challenges.}
Table~\ref{tab:open_challenges} lists the challenges reported in the papers. In total 21 papers (19\%) discussed  challenges. Consequently, the reported challenges do not represent consensus nor importance of the challenges. However, most of the  challenges apply to many other papers; yet, these authors have not explicitly mentioned them. The challenges are organized in five groups. Learning performance challenges primarily relate to timing aspects of learning. Learning effect challenges relate to uncertainties in terms of the effects of using learning in self-adaptation. Domain related challenges are concerned with the characteristics of domains and the transfer of solutions to other problems. Policy related challenges relate to the ability of learning methods to support the principles and rules for decision-making in self-adaptation. Finally, goal related challenges relate to the need for machine learning techniques to support adaptation in practical systems that are characterized by multiple, possible evolving goals. We now zoom in on a few  of the interesting open challenges and outline potential starting points to tackle them.

\begin{table*}[!htbp]
	\centering
	\caption{Open challenges for learning in self-adaptation reported in the papers.}      
	\label{tab:open_challenges}
	\vspace{-5pt}
	\renewcommand{\arraystretch}{1.2}
	\begin{tabu}{|X[0.75,m]X[1.5,m]|X[0.15,c,m]|X[0.6,c,m]|}
		\hline
		\multirow{2}{*}{\shortstack[l]{Learning Performance }} &
		\Tstrut Balance time and accuracy \Bstrut 
		& 3 & \cite{Skalkowski13, Moghadam2018, Pandey2017} \\  \cline{2-4}
		& \Tstrut Handle oscillations in early learning stages  \Bstrut & 1 & \cite{Jamshidi2016} \\ \tabucline[1pt]{-}
		
		\multirow{2}{*}{\shortstack[l]{Learning Effect }} &
		\Tstrut Understand the effect of learning on adaptation decisions over time \Bstrut & 3 & \cite{Chen2018, qian2015rationalism, Zhao2017}\\ \cline{2-4}
		& \Tstrut Guarantees on results of machine learning \Bstrut & 1 & \cite{Quin2019} \\ \tabucline[1pt]{-}
		
		\multirow{6}{*}{\shortstack[l]{Domain-Related }}
		& Handle sudden changes & 2 & \cite{pelaez2016dynamic, qin2014presc} \\ \cline{2-4}
		& \Tstrut Handle open world changes \Bstrut & 1 & \cite{Wan2017} \\ \cline{2-4}
		& \Tstrut Balancing diverse sources of input data 
		\Bstrut & 1 & \cite{Stein2018} \\ \cline{2-4}
		& \Tstrut Extend to other application domains \Bstrut & 1 &  \cite{Sensi16} \\ \cline{2-4}
		& \Tstrut Define similarity measures to transfer to other planning problems \Bstrut & 1 & \cite{Pandey2017}
		\\ \tabucline[1pt]{-} 
		
		\multirow{3}{*}{\shortstack[l]{Policy-Related }} &
		Deal with conflicting policies  & 2 & \cite{pelaez2016dynamic, Azlan15} \\ \cline{2-4}
		& \Tstrut Improve policy evolution speed  \Bstrut & 1 & \cite{Han2015} 
		\\ \tabucline[1pt]{-} 
		
		\multirow{3}{*}{\shortstack[l]{Goal-Related }} & Handle multiple goals & 3 & \cite{Jamshidi2017, Esfahani2013, Quin2019}\\ \cline{2-4}
		& \Tstrut Dynamically define utility function  \Bstrut & 1 &  \cite{Sheikhi18}\\ \hline
	\end{tabu}
	\vspace{8pt}
\end{table*}

An open challenge in machine learning for self-adaptive systems is effect uncertainty~\cite{Chen2018,qian2015rationalism}. Effect uncertainty refers to uncertain effects on the system that may occur when a learner selects a configuration or a plan for adaptation that is applied on the system. Relying on the results of machine learning comes with some degree of (statistical) uncertainty that may affect the decision-making of a self-adaptive system. Detecting and handling this type of uncertainty is an open challenge. Note that this challenge is not specific to machine learning methods. However, we raise it here as many studies have adopted machine learning methods for proactive decision making, where effect uncertainty gets more challenging by the uncertainty introduced by learning methods. 

Learning about open-world changes is another  open challenge in self-adaptive systems. Open world changes have been studied in the field of machine learning under the umbrella of ``lifelong machine learning''~\cite{thrun1995lifelong}, in particular in relation to dealing with new learning tasks. A lifelong machine learner relies on an online learning pipeline that exploits historical knowledge to evaluate and update an existing learner to deal with new tasks. It may be possible to exploit such an approach to support a data-driven self-adaptive system with changes that were not fully anticipated.

In large-scale self-adaptive systems, the feedback loop's monitor component may be distributed over many different nodes (for instance sensor nodes) deployed on the managed system or in the environment. An example is described in~\cite{yamagata2019self} where distributed monitoring components are used in online games that run on a client-server infrastructure. A crucial aspect of the sensed data on the decision-making for adaptation is the impact of heterogeneous sensor data~\cite{Stein2018}, which may dynamically change. Automated weighting of heterogeneous data sources based on the current situation of the system to assure proper decision-making for adaptation is an open challenge. 

One of the characteristic use cases of machine learning is reducing large adaptation spaces to support efficient analysis of different configurations based on model checking at runtime, see for instance~\cite{Quin2019}. An open problem is to understand the impact of the learning process on the results of the model checker as this will affect the guarantees of the decisions made by the feedback loop. Such understanding will not only provide bounds on the expected impact of learning on the guarantees for decision-making in self-adaptive systems, it will also pave the way to dynamically balance the guarantees that are required with the resources that are available to provide them. 

Transfer learning focuses on storing knowledge obtained from solving one problem and applying it to a different but related problem. For example, knowledge gained while learning to recognize anomalies in one type of communication network could then be exploited to recognize anomalies in another type of network. Transfer learning can help self-adaptation by reducing the cost of continuous training and data collection~\cite{Jamshidi2017}. However, this study highlights that transfer learning has rarely been used in self-adaptation so far. 
Inspiration to tackle this open challenges is provided~\cite{Han2015}.

Another important open challenge when using leaning is handling multiple goals. Different approaches exist to deal with multiple goals in self-adaptation, such as utility functions~\cite{Sheikhi18}, \mbox{(semi-)ordered} rules~\cite{Quin2019}, and multi-objective functions~\cite{Esfahani2013}. A key issue of handling multiple goals is balancing time and recourse usage with finding a (close to) optimal solution. Hence, exploring the use of machine learning methods for efficient multi-objective optimization with guarantees on the precision of the results is an open challenge in machine learning for self-adaptation. 

\vspace{0.3cm}
\begin{tcolorbox}[breakable]
	\textbf{Answer to RQ3 - What are open challenges for using machine learning in self-adaptive systems?} Based on the reported limitations and challenges, we identified three broad categories of open challenges. The first category is about quality related challenges. These include the scalability and performance of learning, and the reusability of solutions. The second category is about effect related challenges. Central here are guarantees when learning is applied to support the feedback loop, uncertainties caused by learning and the effects of learning on decision-making. The third category is about design challenges. These include challenges related to the domain at hand, and policy and goal related challenges. 
\end{tcolorbox}

\section{Insights Derived from the Study and Threats to Validity}\label{sec:discussion}

Based on the insights derived from this systematic literature we start this section by outlining an initial design process for applying machine learning in self-adaptive systems. Then, we discuss a number of remarkable observations of the survey that open interesting opportunities for future research. Finally, we discuss threats to validity of the research presented in this paper. 

\subsection{Towards a Design Process for Using Machine Learning in Self-Adaptive Systems}

\begin{figure}[htbp]
	\centering
	\includegraphics[width=1\linewidth]{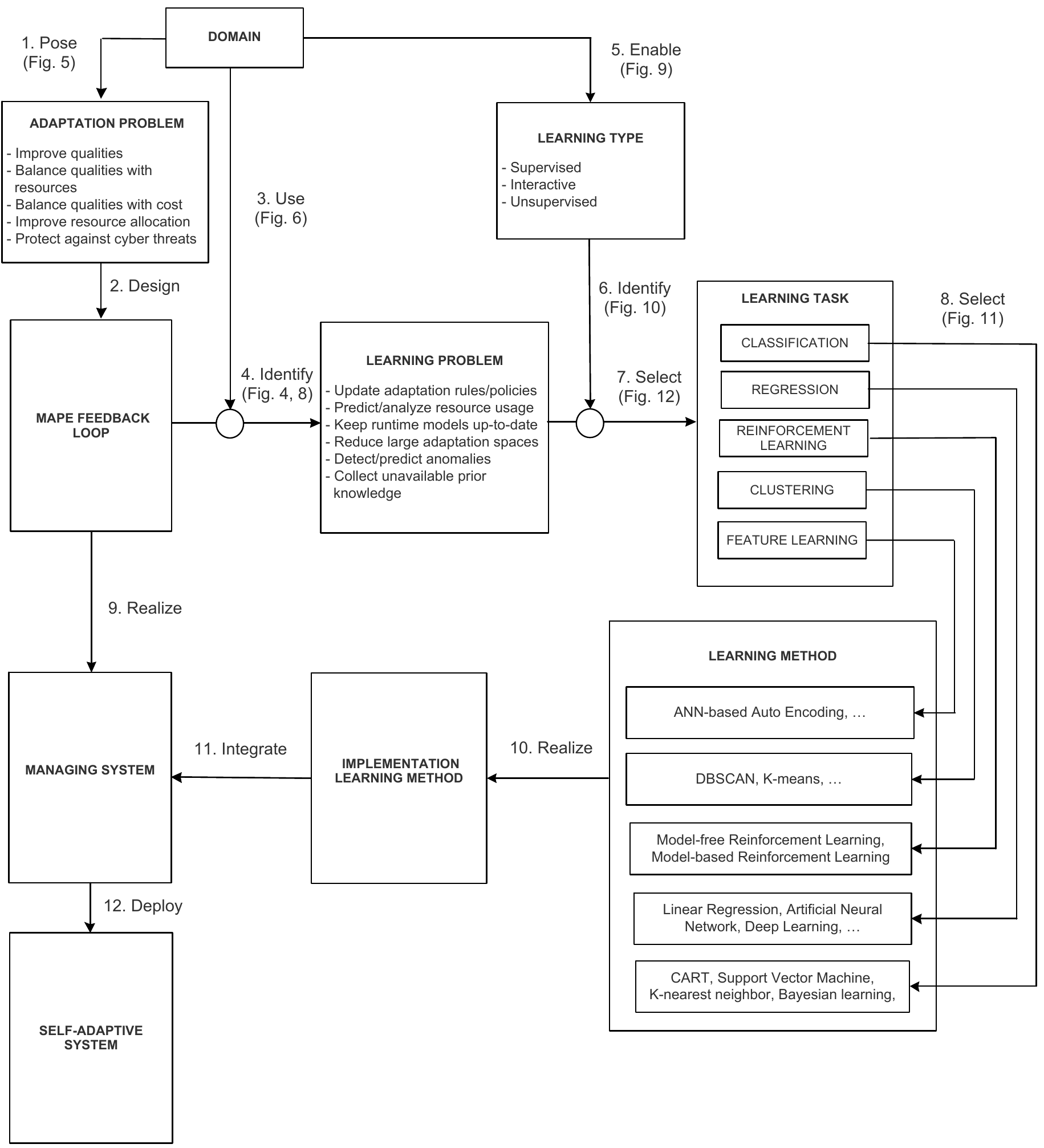}
	\caption{Towards a design process for applying machine learning in self-adaptive systems.}
	\label{fig:process}
\end{figure}

From the results of this review, we present an initial design process that can help guiding designers when applying machine learning in self-adaptive systems that are based on MAPE feedback loops.  While resources exist that support engineers with the design of machine learning techniques in general, see for instance~\cite{2347736.2347755,Julian2016}, to the best of our knowledge, no such design process has been described for self-adaptive systems. 
Figure~\ref{fig:process} shows the different elements of the design process we propose with the conceptual flow of activities between the elements.  

The process starts with defining the adaptation problem that is posed by the domain (\textit{1. Pose}). In this survey, we identified five types of adaptation problems that are solved by MAPE feedback loops supported by learning, see Table\,\ref{tab:adaptation problems}. These types can be instantiated for the problem at hand, supported by the data summarized in Figure\,\ref{fig:adaptation problems vs application domains}. However, the list of adaptation problems can be extended when machine learning is applied to different types of domains and problems.  In the next activity, the MAPE feedback loop of the managing system is designed that aims at solving the adaptation problem (\textit{2. Design}). This means that the designer identifies the knowledge that is maintained by the feedback loop, the functionality that is required to monitor the managed system and its environment, to analyze the runtime data, to plan the actions for adaptation, and to enact these actions. 
Then, the designer uses domain knowledge  (\textit{3. Use}) to identify the learning problem that supports the MAPE feedback loop  (\textit{4. Identify}) (we assume for simplicity here that only a single learning problem needs to be solved by a single learner). The learning problem is delegated to, and solved by a machine learner.  
In this survey, we have identified six different types of learning problems in self-adaptive systems, as shown in Table\,\ref{tab:learning problems}. These types can be instantiated for the problem at hand. To that end, the designer can exploit the data summarized in Figures\,\ref{fig:adaptation vs learning problems}, \ref{fig:learning problems vs application domains} and\,\ref{fig:learning problems vs. MAPE}. Similar to the list of adaptation problems, the list of learning problems can be extended when machine learning is applied to support MAPE feedback loops that deal with new types of adaptation problems in potentially new domains. 

Next the learning task needs to be selected for the learning problem at hand. Besides the characteristics of the learning problem, this choice is determined by the learning type; for instance if no labeled data is available, supervised learning is not an option. The learning type depends both on the characteristics of the domain at hand (\textit{5. Enable} and \textit{6. Identify}) and the realization of the MAPE feedback loop. For the former, the designer can exploit the data summarized in Figures\,\ref{fig:learning_paradigms} and\,\ref{fig:learning_vs_data}. For the mapping of the learning problem to the learning task (\textit{7. Select}), the designer can exploit the data summarized in Figure\,\ref{fig:pvsalg}. 
Next, a concrete learning method is selected to solve the learning task (\textit{8. Select}). This literature review has identified a list of possible learning methods that have been applied to solve different types of learning tasks. The data summarized in Figure\,\ref{fig:ml_methods} supports the designer with selecting a learning method for the learning task at hand. Obviously, new methods can be added to this list as needed.

Finally, the managing system and the learning method are implemented (\textit{9. Realize} and \textit{10. Realize} respectively) and the implementation of the learning method is integrated with the  feedback loop to realize the managing system (\textit{11. Integrate}). This system can then be tested and when accepted, it can be deployed to solve the adaptation problem of the self-adaptive system (\textit{12. Deploy}). 

The proposed process aims at providing a high-level outline of how machine learning can be integrated in the realization of MAPE-based self-adaptive systems.  It is important to note that the flow of activities is conceptual; in practice different activities will be applied iteratively until the system is realized.  Evidently, substantial effort will be required to turn this conceptual idea into a practical engineering process and support it with tools. We put this effort forward as a topic for future research in this area. 

\subsection{Opportunities for Future Research}
\label{subsec:discussion}

The reported limitations of learning methods applied in self-adaptation (Table~\ref{tab:limitations}) as well as the open challenges for this area (Table~\ref{tab:open_challenges}) identify a number of shortcomings of existing learning approaches and highlight demands that may require the use of other or new learning methods to support self-adaptation. Table~\ref{tab:chal_reported} summarizes the themes of the reported challenges.

\begin{table}[htbp]
	\centering
	\caption{Themes of challenges with concrete focus as reported in the papers.}
	\vspace{-5pt}
	\label{tab:chal_reported}
	\begin{tabular}{|p{3cm}|p{10cm}|}
		\hline
		\textbf{Challenge Theme}     & \textbf{Concrete Focus} \\ \hline
		Qualities          &   	Scalability of learning, remove performance penalty \\ \hline
		Uncertainty  & Monitor uncertainty, detect novelty, support open world  \\ \hline
		Goals & Deal with changing goals, conflict of goals, new types of goals  \\ \hline
		Guarantees    &	Ensure quality goals, avoid sub-optimality, support explainability  \\ \hline
		Domain / Design	  &    Deal with parameter tuning, transfer solutions, reusability of solution   \\ \hline    
	\end{tabular}
\end{table}

The themes in Table~\ref{tab:chal_reported} take the stance of the stakeholders of self-adaptive systems, looking from the perspective of the characteristics, demands, and open problems of these systems. We complement this view now with a set of additional opportunities for future research. To that end, we took a step back and explored prospects for advancing the field by looking at opportunities provided by learning methods. In particular, we looked at  machine learning methods that received less or no attention in existing work, and explored how self-adaptation may benefit from further investigation into the use of these learning methods.  Table~\ref{tab:chal_additional} summarizes the opportunities.

\begin{table}[htbp]
	\centering
	\caption{Additional opportunities for future research driven by learning methods.}
	\vspace{-5pt}
	\label{tab:chal_additional}
	\begin{tabular}{|p{4cm}|p{9cm}|}
		\hline
		\textbf{Learning Method}     & \textbf{Concrete Opportunities} \\ \hline
		Unsupervised learning       &   Detecting new structures in complex data, support other learning methods	 \\ \hline
		Active learning  &  Involve stakeholders in decision-making, reduce learning cost, increase speed of learning \\ \hline
		Adversarial learning & Improve rules and policies, detect anomalies  \\ \hline
		Other learning methods    &	Detection of novel phenomena in environment, synchronize execution workflows in complex settings \\ \hline
	\end{tabular}
\end{table}

Only a small fraction (roughly 10\%) of the papers apply unsupervised learning methods. This is remarkable given that one of the key drivers for applying self-adaptation is automating tasks in systems that are subject to uncertainty~\cite{Weyns19}. Since unsupervised learning methods can work independently of external input and do not require labeled data, we observe an interesting opportunity here to further explore unsupervised learning in self-adaptation. Unsupervised learning methods can be used as independent learning techniques to support self-adaptation; one interesting use case is the detection of new structures in complex high-dimensional data~\cite{6482137}. Unsupervised learning methods can also be used to support supervised or even interactive learning methods. A good example here is auto-encoders that have been used to increase the precision of other learning methods by providing a denser representation of data~\cite{xiao2017self, Papamartzivanos2019, Maimo2018}.

On the other hand, most papers apply passive learning. Active learning methods~\cite{settles2009active} interact with the environment or stakeholders to obtain the desired outputs at new data points. This helps to improve the performance of the machine learner by exploring the most informative data. In the context of self-adaptation, applying active learning provides an opportunity to involve stakeholders in the decision-making process, which has been highlighted as a key aspect of establishing trust~\cite{Weyns2018BP}. Active learning can be exploited to reduce the learning cost and increase the convergence speed of learning. It can also be particularly useful to learn updating goals through interaction with stakeholders. In this way, the system can gradually learn essential knowledge of the stakeholder.

Adversarial learning aims at enabling a safe adoption of machine learning techniques in adversarial settings~\cite{45816}. An adversarial machine learner tries to fool a learning model by supplying deceptive input. Adversarial learning can be particularly useful in domains that are sensitive to privacy and security issues, e.g., for signature detection and bio-metric recognition.
Our study shows that only two papers have exploited non adversarial learning in self-adaptation that both adopt a game-theoretical learning approach. This observation opens opportunities to exploit various adversarial learning methods. One example is the use of generative adversarial networks to improve rules and policies in rule- and policy-based systems, or detect new anomalies for self-protection.

The majority of papers (85\%) apply learning to support decision-making in self-adaptation, i.e., the analysis and planning functions. The main use cases are updating and changing rules and policies and predicting and analyzing resource usage. Only a limited number of papers (14\%) apply learning to support monitoring, and here the main use case is keeping runtime models up-to-date.  
Only a single paper applies learning to support execution. This clearly opens opportunities for other use cases. One interesting opportunity is to use learning to support the detection of novel phenomena in the environment that have an effect on the self-adaptive system. Tackling this type of uncertainty is broadly seen as a key challenge in self-adaptive systems~\cite{9196226}. Another opportunity is to exploit learning in the execution of adaptation plans. In complex settings, for instance in large-scale applications with distributed feedback loops, the workflow and synchronisation of adaptation actions is often very difficult to establish manually. Machine learning can then be exploited to learn the best possible execution of the workflow under changing conditions.

\subsection{Threats to Validity}
We list the main threats to validity of this study and the measures we took to mitigate them. 

\vspace{3pt}\noindent 
\textbf{Internal validity}: refers to the extent to which a causal conclusion based on a study is warranted. Potential bias of reviewers is a common validity threat of literature reviews. 
For instance, a reviewer may be biased in the  interpretation of fundamental concepts, i.e., machine learning and self-adaptive system.
To mitigate this risk, we took two measures. First, the three researchers involved in the study defined a protocol before starting the review process to clarify the definition of fundamental concepts and the process to follow. Second, the three researchers were involved in the selection of papers, the data collection and the analysis. A subset of the papers was handled independently by two researchers. The decisions on including or excluding papers and collecting data from the selected papers were based on an agreement between the two researchers. In case of disagreement, a third researcher was consulted, and after discussion, a decision was made in consensus.

\vspace{3pt}\noindent 
\textbf{External validity}:
refers to the generalizability of findings. Applied to this study, this threat is about generalization of the outcome and conclusions of the literature review. By limiting the automatic search to three online libraries, we may have missed some papers. To mitigate this threat, we applied the search string to the main libraries for publishing research in this area. This aligns with other literature reviews. In addition, we crosschecked that established venues for publishing papers in self-adaptation are covered. 
Furthermore, the search string we used may not provide the right coverage of papers. We mitigated this threat by starting the search process with pilot searches to define and tune the search string by collecting data from specific venues via the scientific search engines and comparing the results with manual inspection of the papers of the searched venues. 

\vspace{3pt}\noindent 
\textbf{Construct validity}: refers to the degree to which a study measures what it aims to measure.  Here, the quality of reporting of studies may be a threat as this element affects the validity of the collected data. To anticipate this threat, we extracted data about reporting quality. The analysis of this data shows that the quality of reporting of the papers is of sufficient good quality. This result provides a solid basis to
derive conclusions from extracted data.
Moreover, to mitigate this threat, we excluded all short papers and papers that do not provide a minimum level of assessment. For instance, we excluded~\cite{Frommgen2015} although the topic is relevant for our study, but this is a short paper. Similarly, we excluded~\cite{Sharifloo2016} since this regular paper does not provide a sufficient level of assessment.

\vspace{3pt}\noindent 
\textbf{Reliability}: refers to assuring that the 
research findings can be replicated by another researcher. Here bias of researchers is also a potential validity threat. As explained above, to mitigate this threat, we defined a detailed protocol that provides the necessary guidelines for performing the different steps of the study. Multiple researchers did the paper selection, data extraction and analysis. Another technical threat concerns the methods used to collect papers from the search engines. For example, a search engine may change the operator to select any paper that complies with a query from a star (*) operator to an ``ANY'' operator (ignoring the initial version of the operator). To anticipate this threat, all the review material is available online, enabling a replication of the study.

\section{Conclusion}\label{sec:conclusion}
This literature review aimed at shining a light on the state of the art of using machine learning in self-adaptive systems. The review confirms the rapidly growing research interests in this area. 
We identified six types of problems in self-adaptation that are solved by using machine learning: updating and changing adaptation rules and policies, predicting and analyzing resource usage, keeping runtime models up-to-date, reducing large adaptation spaces, detecting and predicting anomalies, and collecting unavailable prior knowledge. These problems are primarily solved to support analysis and planning in self-adaptation. Supervised and interactive learning dominate, primarily to solve regression, classification and reinforcement learning tasks. The reported limitations and challenges relate to quality properties when learning is used in self-adaptation, the effects of learning on the decision-making, and managing challenging aspects of the domain at hand. 

From the data analysis, we identified an initial process to support designers that want to apply machine learning in the realization of self-adaptive system. We defined an open process that can be extended with new knowledge as we learn more about applying learning in self-adaptive systems. 

Finally, we outlined a number of interesting opportunities for further research in this area, in particular, managing effect uncertainty, dealing with open world changes, dealing with distribution and heterogeneity of data, determining the bounds on guarantees for the adaptation goals implied by the use of machine learning, exploiting transfer learning to related problems, and finally dealing with more complex types of adaptation goals. We hope that the results of this systematic literature review will inspire researchers to tackle these and other problems in this fascinating research area.


\end{document}